%% file: ms.tex
\documentclass{article}
\usepackage{spconf}
\usepackage[hidelinks]{hyperref}

\newcommand\blfootnote[1]{%
  \begingroup
  \renewcommand\thefootnote{}\footnote{#1}%
  \addtocounter{footnote}{-1}%
  \endgroup
}

\input{defpack}
\title{\LARGE \bf
C\MakeLowercase{ommunication}-E\MakeLowercase{fficient} Z\MakeLowercase{eroth}-O\MakeLowercase{rder} D\MakeLowercase{istributed} O\MakeLowercase{nline} O\MakeLowercase{ptimization:} A\MakeLowercase{lgorithm}, T\MakeLowercase{heory}, \MakeLowercase{and} A\MakeLowercase{pplications}}
\name{Ege C. Kaya*, M. Berk Sahin*, and Abolfazl Hashemi\thanks{*The first two authors contributed equally to the manuscript.}}
\address{School of Electrical and Computer Engineering, Purdue University}

\begin{document}

\maketitle

\begin{abstract}
\blfootnote{Ege C. Kaya, M. Berk Sahin, and Abolfazl Hashemi  are with College of Engineering, Purdue University, West Lafayette, IN 47907, USA. Emails: \{\href{mailto:kayae@purdue.edu}{\texttt{kayae}}, \href{mailto:sahinm@purdue.edu}{\texttt{sahinm}}, \href{mailto:abolfazl@purdue.edu}{\texttt{abolfazl}}\}\texttt{@purdue.edu} }This paper focuses on a multi-agent zeroth-order online optimization problem in a federated learning setting for target tracking. The agents only sense their current distances to their targets and aim to maintain a minimum safe distance from each other to prevent collisions. The coordination among the agents and dissemination of collision-prevention information is managed by a central server using the federated learning paradigm. The proposed formulation leads to an instance of distributed online nonconvex optimization problem that is solved via a group of communication-constrained agents. To deal with the communication limitations of the agents, an error feedback-based compression scheme is utilized for agent-to-server communication. The proposed algorithm is analyzed theoretically for the general class of distributed online nonconvex optimization problems. We provide non-asymptotic convergence rates that show the dominant term is independent of the characteristics of the compression scheme. Our theoretical results feature a new approach that employs significantly more relaxed assumptions in comparison to standard literature. The performance of the proposed solution is further analyzed numerically in terms of tracking errors and collisions between agents in two relevant applications.
\end{abstract}


\section{Introduction}\label{section:intro}
As datasets and machine learning (ML) models continue to grow in size and complexity, training ML models increasingly requires carrying out the optimization process across multiple devices. This is often the result of parallel processing needs or the collaboration of multiple participants in the data acquisition and optimization processes. The federated learning (FL) paradigm \cite{FedAvg, privacy}  addresses this by focusing on the latter scenario and training a global model through the cooperation of multiple clients (or agents), managed by a central server. However, FL is typically carried out by a large number of \textit{communication-constrained} agents, making the transmission of model parameters to the central server a potential bottleneck that needs to be addressed for efficient model training.

In online learning (OL), where decisions are made in real-time with limited information/feedback provided to the decision maker, limited communication resources become even a more severe problem.  To address this, first-order FL algorithms like local stochastic gradient descent (SGD) use compression techniques like quantization or sparsification \cite{QSGD, dropout,randk}  to reduce the size of local gradients before transmission, but this causes information loss which may impact the learning performance adversely. 

To counteract this loss in information, an error feedback (EF) mechanism can be added. The EF mechanism works by incorporating the error made by compression in the subsequent steps, so that effectively, each gradient is fully utilized, even if at later stages. Moreover, the EF mechanism theoretically achieves the same rate of convergence as the no-compression case, making compression come at no cost. \cite{errorterm}. 

An additional consideration that we may need to have in practical scenarios is the potentially limited nature of available information. The zeroth-order (ZO) optimization setting presents an example for such limitations. In an optimization problem arising from a real-life scenario, the information to be used in the optimization process may be the sensed values of physical quantities such as sound or light intensity, or relative distance \cite{4599166}. For instance, assuming that sensing agents may only sense current distances to their targets and other nearby agents, we can consider this to be a ZO setting \cite{zeroOrder} as agents do not have access to higher-order information, such as velocity or acceleration.

\textcolor{black}{As an example of a practical scenario combining all of the aforementioned considerations, consider} delivery robots that are loaded from the same region and aim to find their customers. This situation may be viewed as a \textit{source localization} problem with multiple mobile agents. We adopt the terms \textit{agent} and \textit{source} from the literature on this subject in the upcoming discussion. If the customers are also moving, this becomes a target-tracking problem \cite{VALIN2007216, target-tracking}.
\begin{figure}[h]
    \centering
    \includegraphics[width=2in]{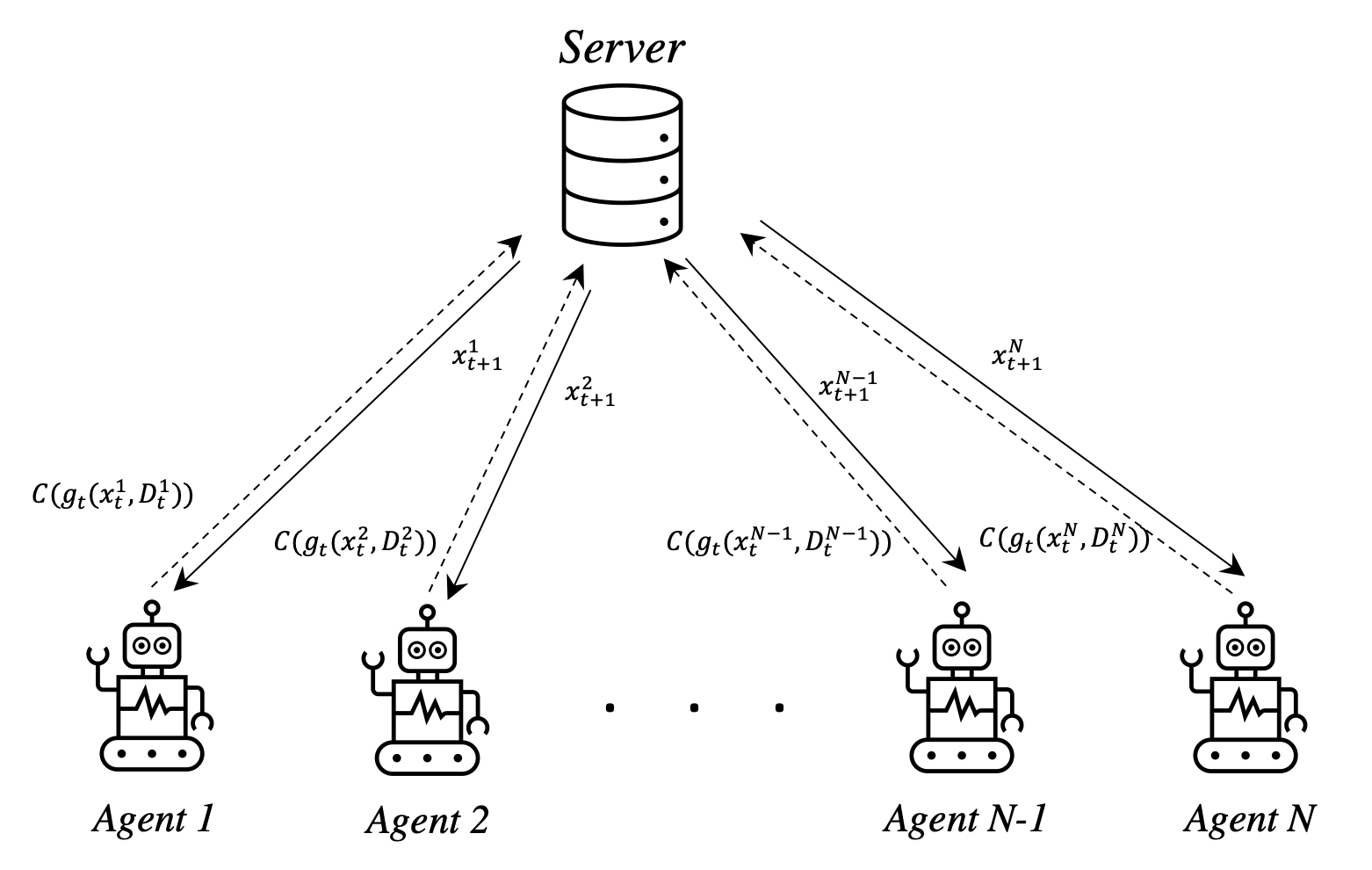}
    \caption{Illustration of agent-server communication. The agents communicate compressed information to the server, whereas the server transmits back the full information. 
    }
    \label{fig:FLillustration}
\end{figure} In a multi-agent setting, collisions between these delivery robots may occur, which can be solved by establishing communication between the agents using the FL framework, somehow incorporating the information of where nearby agents are. Supposing additionally that the robots are only capable of sensing their current distances to their respective targets and to other nearby robots moves our problem into the field of ZO optimization. However, doing so would also result in an online optimization scenario, seeing as the relative locations of the robots with respect to one another would be continually changing, producing a time-varying sequence of optimization problems to solve. Finally, to overcome the inherent communication bottleneck engendered by the online and FL settings, compression schemes may be used along with the EF mechanism. Our novel formulation of this target tracking problem is illustrated and explained in detail in Section \ref{subsec:target-tracking}. 


\subsection{Contribution}\label{section:contribution}
Motivated by the previous problem formulation, the purpose of this work is to find an answer to the central question:

\textit{Is it possible to devise an algorithm for online, distributed non-convex optimization problems with compressed exchange of zeroth-order information, and with provable convergence guarantees for both single-agent and multi-agent settings?}

To address this question, we focus on a general stochastic nonconvex optimization problem, taking into account the following factors: \textbf{i)} access to the stochastic cost function is limited to zeroth-order oracle, meaning only function values at current locations and times are available, \textbf{ii)} due to communication constraints, only compressed or quantized gradients are exchanged between the  agents and the server, \textbf{iii)} multiple agents use zeroth-order information to track their targets, and \textbf{iv)} the objective functions are time-varying in nature, resulting in an online optimization problem.

We prove the existence of a first-order solution in $\mathbb{R}^d$ that is $\xi$-accurate with $T = \mathcal{O}\lp\frac{d\sigma^2ML(\Delta+\Bar{\omega})}{\xi^2}\rp$ in the dominant term, where $\sigma^2$, $L$, $M$, $\Delta$, and $\Bar{\omega}$ denote the variance of stochastic gradients, smoothness constant in Assumption \ref{assump:Lsmooth}, bound constant on the stochastic gradients' second moment in Assumption \ref{assumption:noise}, the difference between averages of loss functions for the first and last iterates, and the summation of drift bounds from Assumption \ref{assump:drift} respectively. Hence the dominant term in the convergence error is not dependent on the compression ratio.
This is achieved while using an EF mechanism and a ZO gradient estimator which 
\textcolor{black}{uses} two function evaluations. In the derivation of this result, we also relax the assumption of bounded second moment commonly found in related literature \cite{errorterm}.
\textcolor{black}{Instead of assuming that the second moment of the stochastic gradients are upper-bounded by a constant term greater than or equal to their variance, we adopt the relaxed assumptıon that it is upper-bounded by the variance plus a term that is proportional to the square of its expected value. In other words, we relax the assumptions on the value of $M$ in Assumption \ref{assumption:noise}, whereas it is commonly assumed in other literature that $M=0$, uniformly. That is, our upper bound depends on the current sample rather than a uniform bound.} Whereas the previous work deals with a single-agent scenario \cite{OPZO}, we examine the effectiveness of the proposed approach in a multi-agent target tracking scenario with limited communication where collision avoidance is of paramount importance. The problem of reducing collisions among agents is addressed by incorporating the FL paradigm and a new regularization term. This task is formulated as an online, distributed nonconvex optimization problem that can be solved by a multi-agent variation of the proposed scheme. Theoretical analysis shows that a $\xi$-accurate first-order solution in $\mathbb{R}^{Nd}$ with $T = 
\mathcal{O}\lp\dfrac{\sigma^2 dMQ\lp\Delta^2+\Bar{\omega}^2\rp+M\lp\sigma^2+Z^4\rp}{\xi^2}\rp$ in the dominant term can be found in a scenario with $N$ agents, where $Z^2$ and $Q$ are constants that arise from Assumption \ref{assump:boundedgrad}, \textcolor{black}{which effectively places a bound on the norm of the gradients of each client in terms of the average of these gradients over all of the clients \cite{decentralizedGD}.} The results of the study are further supported by experimental results 

Our preliminary work on single-agent convergence analysis and experiments was accepted to and will be presented at 2023 IEEE International Conference on Acoustics, Speech, and Signal Processing\cite{OURPAPER}. The current work presents a significantly more thorough analysis of the subject, with an additional part detailing the multi-agent algorithm and its analysis, presented in Theorem 2. The complete proofs of the two theorems are also provided. The experimental section is ameliorated with more descriptive results, and an additional experiment involving an area coverage problem.

\subsection{Related Work}\label{section:related}
\textbf{Communication-efficient FL.} FedAvg is the seminal FL paper in which the central server takes the average of the local gradients transmitted by the clients and distributes the updated parameters to corresponding clients \cite{FedAvg}. The crux of this work is the locality of data, in that data is acquired and trained on locally by a multitude of clients, without ever transporting it to a central server. Several difficulties, including privacy concerns \cite{FEDPriv}, heterogeneity of client data \cite{FedNONIID}, and high communication costs in agent-to-server links \cite{FedCommCost} arise in relation to this paradigm. Variations of FedAvg have been developed to mitigate these problems. For instance, to deal with high communication costs,  \cite{chen2021communication, chen2021decentralized} propose a sparsification algorithm in communication for time-varying decentralized learning and optimization. Reference \cite{AdaptOpt} proposes utilizing adaptive learning rate for aggregation, which is relevant to both the client data heterogeneity and communication efficiency issues. Reference \cite{CommEff} suggests using a novel aggregation technique which first quantizes gradients, and then skips communicating less impactful quantized gradients in favor of reusing previous ones. Reference \cite{das2022faster} proposes using a momentum-based global update at the server, which promotes communication efficiency through variance reduction. Reference \cite{AirComp} proposes a derivative-free federated ZO optimization (FedZO) algorithm, and to improve its communication efficiency over wireless networks, they propose an over-air computation assisted variant. Reference \cite{ZeroHetero} proposes a multiple local update strategy and a decentralized ZO algorithm to improve the communication efficiency and convergence rate in the decentralized FL scenario, in which there is no access to first-order derivatives. Reference \cite{hashemi2021benefits} promotes the use of multiple gossip steps for communication efficiency. Various compression schemes such as Top-k \cite{errorterm}, Rand-k \cite{randk}, Biased and Unbiased Dropout-p \cite{dropout}, Quantized SGD (QSGD) \cite{QSGD}, and their variants/generalizations are used to achieve the communication efficiency of FL algorithms. Compression schemes can be divided into contractive and non-contractive methods. With contractive compression schemes, which are our focus in this work, it is common to introduce an EF mechanism to compensate for the error due to compression by accumulating compression error in memory and adding it back as feedback for subsequent rounds. In \cite{errorterm}, it is shown that such a method used in conjunction with SGD has a comparable rate of convergence to non-compressed SGD. In this study, we relax the assumption of having stochastic-first order oracle with bounded noise required in \cite{errorterm} by meticulously characterizing the impact of such relaxation on convergence. Furthermore, instead of the single-agent case as investigated in \cite{errorterm}, we consider multiple agents with the additional contingency of preventing their collisions, which make the theoretical side more challenging.

\textbf{Multi-agent target tracking.} In our setting, agents are limited to ZO information, since they are assumed to only be able to sense their distance to their targets and other nearby agents. As a result of this consideration, our method is applicable to different practical scenarios such as \cite{EnergyPrice, Bandit}. In these kinds of scenarios, gradients of the loss function can still be estimated by finite differences \cite{ZO} but doing so in a multi-agent setting under communication constraints still remains an open challenge. Reference \cite{OPZO} describes a setup comparable to online optimization employing ZO oracles, applied to a target tracking problem. In that work, the authors focus on the case where there is a single source pursued by a single agent, which we generalize to the multi-agent setting as part of our contribution. We further investigate an effective approach via nonconvex regularization for collision avoidance. The $\lambda$ parameter we refer to as the \textit{regularization parameter} is in essence similar to the penalty and augmented Lagrangian methods used in functional constrained optimization\cite{nocedal_wright_2006, LuLambda}. However, these methods aim to adaptively tune the $\lambda$ parameter on-the-go, which is out of the scope of our work. It should be noted also that this line of research is very relevant to the area of \textit{safe reinforcement learning}, see e.g. \cite{saferl1, saferl2, saferl3, Lu_Zhang_Chen_Başar_Horesh_2021}.

In \cite{AdaptiveBehaviors}, a cooperative, mobile multi-agent source localization problem is tackled via using a distributed algorithm. Compared to our setting, the agents sense first-order information and their neighboring agents benefit from collaboration between agents to avoid collision. Reference \cite{adaptivesource} deals with a source localization problem in a single-agent and single-source setting, where the source is stationary or near-stationary. Reference \cite{TrackingandRegret} studies the problem of OL using ZO information with convex cost functions. They extend the problem out of the conventional Euclidean setting onto Riemannian manifolds. In \cite{Fidan2008GuaranteeingPC} a ZO source localization problem is considered using distance information, where the agents are essentially multiple sensors of known position. References \cite{ZOInFormation} and \cite{Multiagent} deal with an online optimization problem using a decentralized network of multiple agents which have access to ZO information, and propose the usage of local information and information from neighboring agents in the network. In \cite{ZOInFormation}, an iterative algorithm with guarantees is proposed for time-varying online loss functions. The theoretical result there is established by assuming a certain  \textit{bounded drift in time} assumption which is standard in the literature, see, e.g., \cite{OPZO}.

\textcolor{black}{\textbf{Online optimization and online target tracking.} In the general online optimization setting, we focus on literature within or similar to the online convex optimization framework, which we can consider as a sequential decision making game in the presence of time-varying loss functions \cite{OCO}. In \cite{distributed-online}, a distributed online optimization problem with multiple agents is considered, and the local loss function of each agent is convex and time-varying. The authors propose a randomized gradient-free distributed projected gradient descent procedure, where agents estimate the gradient of their local loss functions in a random direction using information from a locally-built ZO oracle. In \cite{continuous-time-varying}, a similar setting is considered, and a multi-agent distributed optimization problem is studied in continuous-time, with time-varying convex loss functions. Reference \cite{OPZO} deals with a setting where zeroth-order oracles are used for optimization in the presence of  time-varying cost functions. Besides the general online optimization setting, there is an abundance of literature focusing on the online optimization aspect of the target tracking problem. A large number of them also involve a swarm of multiple agents working in coordination. Usually, literature on this area tends to consider the problem in the context of unmanned aerial vehicles (UAV), or unmanned surface vehicles (USV). As pointed out in \cite{online-ref-1}, the approaches to the problem may be separated into three broad categories: those using \textit{filtering based, control theory based and machine learning based} approaches. For instance, \cite{online-ref-1} examines the problem within the domain of reinforcement learning by formulating it as a constrained Markov decision process, with application to autonomous target tracking using a swarm of UAVs. The authors provide an algorithm with provable guarantees. In \cite{online-ref-6}, the authors again consider a multi-agent multi-target pursuit evasion scenario, where they propose the usage of a recurrent neural net work for target trajectory prediction, in conjunction with a multi-agent deep deterministic policy gradient formulation for decision making. Reference \cite{online-ref-11} deals with a robust formulation of a similar scenario in the domain of supervised learning, using a game theoretic approach. Reference \cite{online-ref-9} deals with a multi-target following scenario with consideration of external threats. The authors treat the problem as an online path planning problem and adopt a control-theoretic approach. In \cite{online-ref-7}, an online adaptive Kalman filter is ussed in a target tracking problem where the sensor signals of the agents are assumed to have unknown noise statistics, to formulate a solution that is robust to noise. Lastly, \cite{online-ref-12} considers a decentralized control problem involving multiple agents with multiple control objectives, among which target tracking is one. The authors make use of a scheme based on adaptive dynamic programming, and feedback from a critic neural network which approximates the control objectives in online fashion.}

\subsection{Novelty w.r.t. Existing Works}
Our work is focused on a nonconvex online distributed optimization problem with compressed exchange of zeroth-order information, along with the error feedback mechanism. Although these concepts were investigated individually in prior works \cite{errorterm, OPZO, AirComp, ZO, AdaptiveBehaviors} to the best of our knowledge, we are the first to combine them in a single framework and propose an algorithm with its theoretical analysis and convergence guarantee.  

Furthermore, we may compare our theoretical results in Section \ref{section:alg} with the result of the analysis in \cite{zheng2019communicationefficient}, which derives an upper bound for the offline, first-order case with contractive compressors and the error feedback mechanism for the optimization of a smooth, nonconvex function in the FL paradigm. Their result establishes, in this setting, an iteration complexity of $\mathcal{O}(1/N\xi^2)$ to produce a $\xi-$accurate first-order solution. Our result agrees with this result on the $\xi-$dependency. However, the analysis in this result is obtained in an offline setting with access to the first-order derivatives and is hence able to derive a convergence rate that is inversely proportional to the number of agents $N$. As opposed to that setting, we consider a more challenging online setting in which agents lack access to first-order derivatives but have access to finite differences. Thus, the convergence rate we establish is independent of $N$. Moreover, reference \cite{zheng2019communicationefficient} assumes a uniform bound on the second moment of the gradients. We adopt a more relaxed version of this assumption, which does not assume a uniform bound on the second moment (Assumption \ref{assumption:noise}), and this makes our analysis more involved. We make a similar relaxation of a standard assumption used in distributed optimization \cite{linearSpeedup, scaffold, tighter, unified, adaptiveFedAvg, tacklingHetero, slowmo} in the same vein, by lifting the assumption of a uniform bound (Assumption \ref{assump:boundedgrad}).

The rest of the paper is organized as follows: In Section \ref{section2}, we present the related background on stochastic gradient descent in zeroth-order oracle setting and necessary assumptions for our theoretical analysis. In Section \ref{section:alg}, we propose the \nameref{alg:ef-zo-sgd} and \nameref{alg:fed-ef-zo-sgd} algorithms and present two theorems for their convergence along with sketches of proofs. Experimental results for two different settings are presented in Section \ref{sec:experimental-results} followed by the conclusion in Section \ref{conclusion}. The complete proofs of the theorems along with the statements of relevant lemmas are presented in the Appendix. 

\section{Preliminaries and Background}\label{section2}
We start by providing a description of the problem in the single-agent setting. We deal with a sequence of time-varying  optimization problems: $\min_{x\in\mathbb{R}^d}\ell_t(x)$, $t \in \mathbb{Z}^+$. Each $\ell_t : \mathbb{R}^d \to \mathbb{R}$ is a continuous loss function and $\ell_t(x) := \mathbb{E}_z[\Tilde{\ell}_t(x)]$. We denote  $\Tilde{\ell}_t(x):=\ell_t(x,z)$ where $z$ is a random variable representing data points coming from the unknown distribution $P_z$, so $z\sim P_z$. In our application, the target tracking problem, it is the position vector of targets. We aim to find a sequence of solutions $\{x_t\}_{t=1}^T$ such that $\frac{1}{T}\sum_{t=1}^T\|\nabla\ell_t(x_t)\|^2\leq \xi$ for some  small $\xi>0$. Suppose that, at time $t$, we have somehow generated a (possibly non-optimal) solution $x_t$ to the problem $\min_{x\in\mathbb{R}^d}\ell_t(x)$. As we are motivated by online and time-critical missions, we would like to generate a solution $x_{t+1}$ to the problem $\min_{x\in\mathbb{R}^d}\ell_{t+1}(x)$ applying a simple update rule which is similar to SGD to $x_t$: 
\begin{equation}\label{eq:update}
    x_{t+1} = x_t - \eta_t \nabla\Tilde{\ell}_t(x_t),
\end{equation} where $\eta_t$ is the step size or learning rate adopted at time $t$. As discussed in Section \ref{section:intro},  we cannot directly apply such an update since we are in the ZO setting, that is, we only have access to evaluation of $\Tilde{\ell}_t$ and not to its gradient or stochastic gradient. To overcome this limitation, we resort to a ZO estimator of the gradient:
\begin{equation}\label{appg}
    \Tilde{g}_{\mu,t}(x_t) \vcentcolon= \frac{\Tilde{\ell}_{t}(x_t + \mu u_t) - \Tilde{\ell}_t(x_t)}{\mu}u_t,
\end{equation} where $\mu \in \mathbb{R}$ is the so-called smoothing parameter, and each $u_t \sim\mathcal{N}(0, I_d)$. Note that $\Tilde{g}_{\mu, t}$ can be thought of as an approximation to the stochastic gradient of a Gaussian smoothing of $\Tilde{\ell}_t$, i.e., $\Tilde{\ell}_{\mu, t}(x)\vcentcolon= \mathbb{E}_u[\Tilde{\ell}_t(x+\mu u)]$. A final modification to the update rule arises due to the aforementioned communication constraints. We apply compression to the ZO estimator and use the resulting quantity in the update rule. To mitigate the negative effect of compression on the convergence of the method, we employ the error feedback mechanism. Essentially, this serves in each time step to partially recover information discarded in the previous compression steps. The details of our approach may be seen in \nameref{alg:ef-zo-sgd}.

In the multi-agent setting, we generalize the problem as follows: There are now $N$ sequences of continuous loss functions where $t \in \mathbb{Z}^+$ and each $\ell_t : \mathbb{R}^{Nd} \to \mathbb{R}$, which we denote $\ell^1_t, \ldots, \ell^N_t,$ belonging to agents $1$ through $N$. Similar to the previous part, $\ell^i_t(x) := \mathbb{E}_z[\Tilde{\ell}^i_t(x)]$,  $\Tilde{\ell}^i_t(x):=\ell^i_t(x,z)$ and $z\sim P_z$. We name these the \textit{local loss functions}, since they represent the loss of each specific agent. The objective is to find a sequence of solutions $\{x^{1:N}_t\}_{t=1}^T \subset \mathbb{R}^{Nd}$ that minimizes the \textit{global loss function} $\Bar{\Tilde{\ell}}_t = \frac{1}{N}\sum_{i=1}^N \Tilde{\ell}^i_t.$ Akin to the single-agent setting, each agent computes a compressed version of the ZO estimator, corrected to some extent by feedback of the error generated due to compression in the previous steps. The result of this computation is then transmitted to the central server, where they are aggregated and used to update the locations of each agent. The full algorithm entailed by this approach can be seen in \nameref{alg:fed-ef-zo-sgd}.

Next, we state the assumptions adopted in the forthcoming analyses of the single- and multi-agent settings.
\begin{assumption}(Unbiased Stochastic Zeroth-Order Oracle)\label{assump:zo}
For any $t\in \mathbb{Z}^+$, $i \in \{1, \ldots, N\}$ and $x\in \mathbb{R}^d$, we have
\begin{equation}
    \mathbb{E}_{z}\left[\Tilde{\ell}^i_t(x) \right] = \ell^i_t(x).
\end{equation}
\end{assumption}
Although we do not explicitly utilize the stochastic gradient $\nabla\tilde{\ell}_t$ in the forthcoming algorithm, our analysis still requires a certain regulatory assumption on it. 
\begin{assumption}(Bounded Stochastic Gradients)\label{assumption:noise}
For any $t\in \mathbb{Z}^+$, $i \in \{1, \ldots, N\}$ and $x \in \mathbb{R}^d$, there exist $\sigma, M > 0$ such that
\begin{equation}
    \mathbb{E}_z\left[\lVert \nabla\Tilde{\ell}^i_t(x)\rVert^2\right] \leq \sigma^2 + M\lVert \nabla\ell^i_t(x) \rVert^2.
\end{equation}
\end{assumption} 
We note that this assumption is significantly more relaxed compared to the assumption typically used in stochastic optimization \cite{lan2020first} and EF-based compression \cite{errorterm}. In particular, \cite{errorterm} requires $M = 0$ which effectively imposes a uniform bound on the gradient of $\ell_t$. As part of our contribution, we carry out the analysis under the relaxed assumption stated above.

\begin{assumption}(L-smoothness)\label{assump:Lsmooth}
Each $\Tilde{\ell}^i_t(x)$ is continuously differentiable and L-smooth over $x$ on $\mathbb{R}^d$, that is, there exists an $L \geq 0$ such that for all $x, y \in \mathbb{R}^d$, $t\in \mathbb{Z}^+$ and $i \in \{1, \ldots, N\}$, we have 
\begin{equation}
\lVert \nabla \Tilde{\ell}^i_t(x) - \nabla \Tilde{\ell}^i_t(y) \rVert \leq L \lVert x - y \rVert.
\end{equation}
We denote this by $\Tilde{\ell}^i_t(x) \in C^{1,1}_L(\mathbb{R}^d)$. Note that this assumption implies $\ell^{i}_t(x)\in C^{1,1}_{L}(\mathbb{R}^d)$.
\end{assumption}
\begin{assumption}(Bounded Drift in Time)\label{assump:drift}
There exist $N$ bounded sequences $\{\omega^1_t\}_{t=1}^T, \ldots, \{\omega^N_t\}_{t=1}^T$ such that for all $t\in \mathbb{Z}^+$ and $i \in \{1, \ldots, N\}$, $\lvert \ell^i_t(x) - \ell^i_{t+1}(x)\rvert \leq \omega^i_t$ for any $x\in \mathbb{R}^d$. Note that in the case where $\ell^i_{t+1} = \ell^i_{t}$, this assumption holds with $\omega^i_t = 0$.
\end{assumption}
Assumption \ref{assump:drift} is standard in the literature on time-varying optimization \cite{OPZO,simonetto2020time}. \textcolor{black}{Since we work in the online optimization setting where our loss function is time-varying, this assumption upper-bounds the change in the loss function uniformly with a different constant value at each time step.}

The next assumption has to do with the aforementioned compression of the gradient estimator $g_{\mu, t}$. We assume that the schemes used for the compression satisfy the following assumption. 
\begin{assumption}(Contractive Compression\cite{errorterm})\label{assump:contractive} The compression function $\mathcal{C}$ is a contraction mapping, that is,
\begin{equation}
    \mathbb{E}_{\mathcal{C}}\left[\lVert\mathcal{C}(x) - x\rVert^2 \mid x\right] \leq \left(1-\delta\right)\lVert x\rVert^2
\end{equation} for all $x\in\mathbb{R}^d$ where $0<\delta\leq 1$, and the expectation is over the randomness generated by compression $\mathcal{C}$. 
\end{assumption}
\textcolor{black}{One can see that $\delta$ effectively controls the scale of the compression. $\delta=1$ corresponds to the case of no compression and the amount of compression increases as $\delta\to0.$}

The compression operators we use in the numerical experiments are as follows:

\begin{itemize}
    \item $\operatorname{top}_k$: We fix a parameter $k \in \{0, \ldots, d\}$. $\operatorname{top}_k: \mathbb{R}^d \rightarrow \mathbb{R}^d$ is defined as: 
    \begin{equation}\label{topK}
        (\operatorname{top}_k(x))_i \vcentcolon= \begin{cases} (x)_{\pi(i)} & i \leq k, \\
        0 & \text{otherwise}.
        \end{cases}
    \end{equation} where $\pi(i)$ is a permutation of $\{1, \ldots, d\}$ such that $(\lvert x\rvert)_{\pi(i)} \geq (\lvert x\rvert)_{\pi(i+1)}$ for every $i \in \{1, \ldots, d-1 \}$ \cite{randk}. In other words, $\operatorname{top}_k$ preserves the $k$ elements of $x$ that are largest in magnitude, and assigns $0$ to the rest.
    \item $\operatorname{rand}_k$: We fix a parameter $k \in \{0, \ldots, d\}$. $\operatorname{rand}_k: \mathbb{R}^d \times \Omega_k \rightarrow \mathbb{R}^d$ is defined as:
    \begin{equation}\label{randK}
        (\operatorname{rand}_k(x, \omega_0))_i \vcentcolon= \begin{cases} x_i & i \in \omega_0, \\
        0 & \text{otherwise}.
        \end{cases}
    \end{equation} where $\Omega_k = \{\omega : \omega \subseteq \{1, \ldots, d\}, \lvert \omega \rvert = k\}$ and $\omega_0$ is chosen uniformly at random from $\Omega_k$ \cite{randk}. In other words, $\operatorname{rand}_k$ preserves $k$ random elements of $x$, and assigns $0$ to the rest.
    \item $\operatorname{dropout-b}_p$: We fix a parameter $p \in [0, 1]$. $\operatorname{dropout-b}_p: \mathbb{R}^d \rightarrow \mathbb{R}^d$ is defined as:
    \begin{equation}\label{dropout-b}
        (\operatorname{dropout-b}_p(x))_i \vcentcolon= \begin{cases}
        (x)_i & u_i \leq p, \\
        0 & \text{otherwise}.
        \end{cases}
    \end{equation}where each $u_i \sim U[0, 1]$. Note that $\operatorname{dropout-b}_p(x)$ is a biased estimator of $x$. 
    \item $\operatorname{dropout-u}_p$: We fix a parameter $p \in [0, 1]$. $\operatorname{dropout-u}_p: \mathbb{R}^d \rightarrow \mathbb{R}^d$ is defined as:
    \begin{equation}\label{dropout-b}
        (\operatorname{dropout-u}_p(x))_i \vcentcolon= \begin{cases}
        \frac{1}{p}(x)_i & u_i \leq p, \\
        0 & \text{otherwise}.
        \end{cases}
    \end{equation}where each $u_i \sim U[0, 1]$. Note that $\operatorname{dropout-u}_p(x)$ is an unbiased estimator of $x$. 
    \item $\operatorname{qsgd}_b$: We fix a parameter $b \in \mathbb{N}$ and perform $b$-bit random quantization (where $2^b$ is the quantization level):
    \begin{equation}
        \operatorname{qsgd}_b(x) = \dfrac{\operatorname{sign}(x)\lVert x\rVert_2}{2^bw}\left[2^b\dfrac{|x|}{\lVert x\rVert_2}+u\right]
    \end{equation} where $w=1+\min(\sqrt{d}/2^b, d/2^{2b})$, $u\sim (U[0,1])^d$, and $\operatorname{qsgd}_b(0)=0$ \cite{QSGD}.
\end{itemize}

It is worth noting that all of these compression schemes respect Assumption \ref{assump:contractive}, with the sole exception of $\operatorname{dropout-u}_p$.

Our final assumption concerns only the analysis of the multi-agent case:

\begin{assumption}(Bounded Gradients)\label{assump:boundedgrad}
For any $x^{1:N}_t\in \mathbb{R}^{Nd}$, there exist $Z,Q > 0$ such that 
\begin{equation}
    \mathbb{E}_z\left[\lVert\nabla\ell^i_t(x^{1:N}_t)\rVert^2\right] \leq Z^2 + Q\lVert\nabla\Bar{\ell}_t(x^{1:N}_t)\rVert^2
\end{equation} for all $i \in \{1, \ldots, N\},$ where $\nabla\Bar{\ell}_t(x^{1:N}_t)=\frac{1}{N}\sum_{i=1}^N \nabla\ell^i_t(x^{1:N}_t).$
\end{assumption}
We note that this is a \textcolor{black}{relaxation of the} standard assumption capturing the effect of data heterogeneity, commonly employed in the analyses of decentralized optimization algorithms \cite{canDecent, decentralizedGD, stochGradPush} and in the analysis of FedAvg-like methods in particular \cite{linearSpeedup, scaffold, tighter, unified, adaptiveFedAvg, tacklingHetero, slowmo}. \textcolor{black}{The standard assumption poses a uniform bound: $\mathbb{E}_{z_{1:T}}\lb\lVert \nabla\ell^i_t(x^{1:N}_t) - \nabla\Bar{\ell}_t(x^{1:N}_t)\rVert^2\rb\leq Z^2$. In \cite{unreasonable}, it is argued that this form usually holds in practice, and may even be considered too pessimistic. However, one can easily come up with a counterexample where it does not, e.g., with $\ell^i_t(x) = (ix)^2$ for all $t \in \mathbb{Z}^+.$ We note that this relaxation of the assumption is akin to the one adopted with Assumption \ref{assumption:noise}.}

\section{Proposed Method}\label{section:alg}

\textcolor{black}{In this section, we present our \nameref{alg:ef-zo-sgd} and \nameref{alg:fed-ef-zo-sgd} algorithms along with their convergence results and provide sketches of the proofs for these results. The complete proofs may be found in Appendix.}

\subsection{EF-ZO-SGD}
We now present \textsc{\nameref{alg:ef-zo-sgd}}, an algorithm which uses compression along with the EF mechanism in addition to the ZO estimator in \eqref{appg} to achieve a communication-efficient method of approaching the presented problem in the single-agent scenario. The complete algorithm is demonstrated in \nameref{alg:ef-zo-sgd}. Given an initial solution $x_0 \in \mathbb{R}^d$, which for our problem represents the initial position of the agent, the algorithm works iteratively to construct subsequent solutions to the sequence of optimization problems. It first samples a random vector in $\mathbb{R}^d$ from the standard Gaussian distribution and uses this to construct a ZO estimator to the gradient (steps 3 and 4). Then, the error feedback vector, which keeps track of information discarded during compression in previous communication rounds (step 7) is added to this ZO estimator to produce the augmented estimator (step 5). In this manner, information previously lost to compression is re-utilized. The augmented estimator is the quantity used in the update rule to produce the subsequent solution (step 6), and it is further used to update the error feedback vector (step 7). This process is repeated for $t=1,...,T$ to produce solutions to all terms of the sequence of optimization problems.

\begin{algorithm}[]
\caption{EF-ZO-SGD}
\label{alg:ef-zo-sgd}
\hspace*{\algorithmicindent}\textbf{Input:} Number of time steps $T \in \mathbb{Z}^+$, smoothing parameter $\mu \in \mathbb{R}$, initial agent position $x_0 \in \mathbb{R}^d$, learning rate $\eta \in \mathbb{R}$, sequence of target positions $\{z_t\}_{t=1}^T \subset \mathbb{R}^d.$ \\
\hspace*{\algorithmicindent}\textbf{Output:} Sequence of optimal agent positions $\{x_t\}_{t=1}^T \subset \mathbb{R}^d.$ 
\begin{algorithmic}[1]
\STATE $e_0 = 0$
\FOR{$t=1,\ldots,T$}
\STATE $u_t\sim\mathcal{N}(0, I_{d})$
\STATE $\Tilde{g}_{\mu, t}(x_t)=\dfrac{\Tilde{\ell}_t(x_t + \mu u_t) - \Tilde{\ell}_t(x_t)}{\mu}u_t$
\STATE $p_t = \Tilde{g}_{\mu, t}(x_t) + e_t$
\STATE $x_{t+1} = x_{t} - \eta \mathcal{C}(p_t)$
\STATE $e_{t+1} = p_t - \mathcal{C}(p_t)$
\ENDFOR
\end{algorithmic}
\end{algorithm} 
The convergence properties of \nameref{alg:ef-zo-sgd} are analyzed next. For the convergence of \nameref{alg:ef-zo-sgd} in a single-agent setting, we establish Theorem \ref{theorem:single}.

Note that although the \nameref{alg:ef-zo-sgd} algorithm can be thought of as a SGD-type scheme, the analysis -- due the interaction of EF and ZO estimation -- is involved. In the proof, we leverage a new \textit{intertwined perturbation analysis}, wherein we analyze the convergence of a virtual solution sequence to the smoothed functions $\ell_{\mu, t}$ and tie that to the performance of the real iterates $x_t$ to $\ell_t$, while utilizing the relaxed bounded stochastic gradient assumption.

\begin{theorem}\label{theorem:single}Suppose Assumptions \ref{assump:zo}--\ref{assumption:noise} hold. Consider \nameref{alg:ef-zo-sgd} algorithm. Then, if
$\eta = \dfrac{1}{\sigma\sqrt{(d+4)MTL}}$ and $\mu=\dfrac{1}{(d+4)\sqrt{T}},$
it holds that
\begin{equation}\label{equ22}
\begin{split}
    &\frac{1}{T}\sum_{t=1}^T\mathbb{E}\lb\lVert\nabla\ell_t(x_t)\rVert^2\rb \leq \frac{8\Delta\sigma(d+4)^{\frac{1}{2}}M^{\frac{1}{2}}L^{\frac{1}{2}}}{T^{\frac{1}{2}}} \\ &+ \frac{8\sigma dL^\frac{3}{2}M^\frac{1}{2}}{T^\frac{3}{2}(d+3)^\frac{3}{2}}+ \frac{2(d+6)^\frac{3}{2}L^\frac{5}{2}}{\sigma (d+4)^\frac{5}{2}T^\frac{3}{2}M^\frac{1}{2}} + \frac{8\sigma(d+4)^\frac{1}{2}L^\frac{1}{2}}{M^\frac{1}{2}T^\frac{1}{2}} \\&+ \frac{(d+3)^3L^2}{(d+2)^2T} + \frac{32L}{\delta^2\sigma^2MT} + \frac{8(d+6)^3L^3}{\delta^2\sigma^2(d+4)^3MT^2} \\&+ \frac{8\Bar{\omega}\sigma(d+4)^\frac{1}{2}M^\frac{1}{2}L^\frac{1}{2}}{T^\frac{1}{2}},
\end{split}
\end{equation} where $x_{T+1}^*\in\argmin_{x\in \mathbb{R}^d} \ell_{T+1}(x),$  $\Delta = \ell_1(x_1)-\ell_{T+1}(x_{T+1}^*)$, $\Bar{\omega}=\sum_{t=1}^T\omega_t,$ and $\E[$\makebox[2ex]{$\cdot$}$]$ denotes $\E_{z_{1:T}}[$\makebox[2ex]{$\cdot$}$]$.
Furthermore, the number of time steps $T$ to obtain a $\xi$-accurate first order solution is 
\begin{equation}\label{eq:complexity1}
    T = \mathcal{O}\left(\frac{d\sigma^2 L\Delta M}{\xi^2} + \frac{dL\Delta}{\delta^2\xi} + \frac{\Bar{\omega}\sigma^2dML}{\xi^2}\right).
\end{equation}
\end{theorem}
\begin{proof}[Sketch of proof]
We begin by defining the perturbed quantity $\Tilde{x}_{t} := x_t - \eta e_t.$ Then, using assumptions \ref{assump:Lsmooth} and \ref{assump:drift}, we obtain the inequality
\begin{equation}\label{boundeddrift-pf}
\begin{split}
    \ell_{\mu, t+1}(\Tilde{x}_{t+1})\leq \ell_{\mu, t}(\Tilde{x}_t) - \eta\langle \Tilde{g}_{\mu, t}(x_t), \nabla\ell_{\mu, t}(\Tilde{x}_t)\rangle \\ + \frac{L\eta^2}{2}\lVert \Tilde{g}_{\mu, t}(x_t)\rVert^2 + \omega_t
.
\end{split}
\end{equation}
Taking expectations and performing algebraic manipulations produce the main inequality with the four terms:
\begin{equation}
\begin{split}
    \underbrace{\frac{\eta}{2}\lVert\nabla\ell_{\mu, t}(x_t)\rVert^2}_\text{Term I} \leq \underbrace{\left[\ell_{\mu, t}(\Tilde{x}_t)-\ell_{\mu, t+1}(\Tilde{x}_{t+1})\right]}_\text{Term II} + \\\underbrace{\frac{L\eta^2}{2}\mathbb{E}_{u_t,z_t}\left[\lVert \Tilde{g}_{\mu, t}(x_t)\rVert^2\right]}_\text{Term III} + \underbrace{\frac{L^2\eta^3}{2}\lVert e_t\rVert^2}_\text{Term IV} + \omega_t.   
\end{split}
\end{equation}
We can upper-bound Term II by means of a telescoping sum. Then, using assumptions \ref{assump:contractive} and \ref{assumption:noise}, Term I can be lower-bounded and Terms III and IV can be upper-bounded by quantities involving $\mathbb{E}_{z_{1:T}}[\lVert \nabla\ell_t(x_t)\rVert^2].$ Rearranging this, inserting the values for $\eta$ and $\mu$ and introducing $\xi$ to obtain an expression for the time complexity lead directly to the result. The complete proof may be found in the Appendix.
\end{proof}
We further note that \eqref{eq:complexity1} demonstrates that the dominant term in the complexity is independent of the compression parameter $\delta$. Therefore, for long sequences of time-varying optimization problems where $T$ is very large, the contribution of compression to the convergence error is negligible. Also notable is the fact that the complexity scales with dimension $d$. While this dependence is undesirable, in the worst case, it is unavoidable even without compression as shown in \cite{duchi2015optimal}.

We may discuss the implication of our results to the setting of learning parameters of an overparameterized model, e.g., a deep learning predictor. It has been argued, see, e.g. \cite{vaswani2019fast,acharya2022robust}, such models typically satisfy a so-called \textit{strong growth condition} which implies $\sigma = 0$ in Assumption \ref{assumption:noise}. That is, as the EF-ZO-SGD algorithm converges to a stationary solution, it enters into a virtuous cycle wherein the noise in the stochastic gradient reduces. As our analysis demonstrates, in such settings we can modify $\eta$ and $\mu$ accordingly (in particular set $\eta$ independent of T) to improve the complexity of the proposed algorithm to $T = \mathcal{O}(\frac{1}{\xi})$.
\begin{algorithm}[t]
\caption{FED-EF-ZO-SGD}
\label{alg:fed-ef-zo-sgd}
\hspace*{\algorithmicindent}\textbf{Input:} Number of time steps $T \in \mathbb{Z}^+$, number of agents $N \in \mathbb{Z}^+$, smoothing parameter $\mu \in \mathbb{R}$, initial agent positions $x_0^{1:N} \in \mathbb{R}^{Nd}$, learning rate $\eta \in \mathbb{R}$, sequence of target positions $\left\{z^{1:N}_t\right\}_{t=1}^T \subset \mathbb{R}^{Nd}.$ \\
\hspace*{\algorithmicindent}\textbf{Output:} Sequence of optimal target positions $\left\{x^{1:N}_t\right\}_{t=1}^T \subset \mathbb{R}^{Nd}.$
\begin{algorithmic}[1]
\FOR{$i=1,\ldots,N$}
\STATE $e_0^i = 0$
\ENDFOR
\FOR{$t=1,\ldots,T$}
\STATEx \textit{Runs on each agent:}
\FOR{$i=1, \ldots, N$} 
\STATE $u_t^i\sim\mathcal{N}(0, I_{Nd})$
\STATE $\Tilde{g}^i_{\mu, t}(x^{1:N}_t)=\dfrac{\Tilde{\ell}^i_t(x^{1:N}_t + \mu u^i_t) - \Tilde{\ell}^i_t(x^{1:N}_t)}{\mu}u^i_t$
\STATE $p^i_t = \Tilde{g}^i_{\mu, t}(x^{1:N}_t) + e^i_t$
\STATE $e^i_{t+1} = p^i_t - \mathcal{C}(p^i_t)$
\STATE $\operatorname{transmit\_to\_server}\left(\mathcal{C}(p^i_t)\right)$
\ENDFOR
\STATEx \textit{Runs on the server:}
\STATE $\mathcal{G}_t = \frac{1}{N}\sum_{i=1}^N \mathcal{C}(p^i_t)$ 
\STATE $x^{1:N}_{t+1} = x^{1:N}_{t} - \eta\mathcal{G}_t$
\STATE $\operatorname{transmit\_to\_clients}\left(x^{1:N}_{t+1}\right)$
\ENDFOR

\end{algorithmic}
\end{algorithm}

\subsection{FED-EF-ZO-SGD}
\textcolor{black}{\nameref{alg:fed-ef-zo-sgd} algorithm is a generalization of \nameref{alg:ef-zo-sgd} to multi-agent and multi-target setting. In addition to the compression, EF mechanism, and ZO estimator, agents are coordinated with the central server and their compressed gradients are averaged in the server as in \cite{FedAvg}. The complete algorithm is shown in \nameref{alg:fed-ef-zo-sgd}. Given an initial solution $x_0^{1:N}\in\mathbb{R}^{Nd}$, which in our problem represents the concatenation of the initial position of the agents, the \nameref{alg:fed-ef-zo-sgd} algorithm works iteratively on both the agent side and the server side to generate the consecutive solutions to the sequence of optimization problems. The agent side is similar to \nameref{alg:ef-zo-sgd} except for the content of the solution vectors. In our setting, without loss of generality, we consider agents which can sense the position of nearby agents called "neighbors" and merge the position vectors with their current position to obtain $x_t^{1:N}$. Entries that correspond to the other agents which are not neighbors are set to 0. The same algorithm can be implemented for the agents having no knowledge of the nearby agents' positions. For every agent, the algorithm first samples a random vector in $\mathbb{R}^{Nd}$ from the standard Gaussian distribution and the entries that do not correspond to $i^{th}$ agent's position are set to 0 (step 6). Thus, only $i^{th}$ agents position vector is perturbed to approximate the noisy gradient with finite differences (step 7). Steps 8 and 9 are the same as \nameref{alg:ef-zo-sgd}. Lastly, each agent sends its compressed augmented estimator to the central server. After the server collects all the estimators from every agent, it takes their average (step 12). Then, this average is used in the update (step 13) and the new positions are transmitted to the agents. This procedure is followed for $t=1,..., T$ to produce solutions to all terms of the sequence of optimization problems. }

Now, we proceed with the analysis extended to the multi-agent case, which involves \nameref{alg:fed-ef-zo-sgd}. We state the following theorem:
\begin{theorem}\label{theorem:multi}Suppose Assumptions \ref{assump:zo}--\ref{assump:boundedgrad} hold. Consider \nameref{alg:fed-ef-zo-sgd} algorithm. Then, if
$\eta = \dfrac{1}{\sigma\sqrt{(d+4)MQTL}}$ and   $\mu=\dfrac{1}{(d+4)\sqrt{T}},$ it holds that
\begin{equation}
    \begin{split}
        &\frac{1}{T}\sum_{t=1}^T\mathbb{E}\lb\lVert\nabla\Bar{\ell}_t(x^{1:N}_t)\rVert^2\rb \leq \dfrac{8\Delta\sigma\lp d+4\rp^{\frac{1}{2}}M^{\frac{1}{2}}Q^{\frac{1}{2}}L^{\frac{1}{2}}}{T^{\frac{1}{2}}} 
        \\&+ \dfrac{8L^{\frac{3}{2}}d\sigma M^{\frac{1}{2}}Q^{\frac{1}{2}}}{\lp d+4\rp^{\frac{3}{2}}T^{\frac{3}{2}}}
        + \dfrac{8L^{\frac{1}{2}}\lp d+4\rp^{\frac{1}{2}}M^{\frac{1}{2}}Z^2}{\sigma Q^{\frac{1}{2}} T^{\frac{1}{2}}}
        \\&+ \dfrac{8L^{\frac{1}{2}}\lp d+4\rp^{\frac{1}{2}}\sigma}{M^{\frac{1}{2}} Q^{\frac{1}{2}} T^{\frac{1}{2}}}
        + \dfrac{2L^{\frac{5}{2}}\lp d+6 \rp^3}{\lp d+4 \rp^{\frac{3}{2}}T^{\frac{3}{2}}\sigma M^{\frac{1}{2}} Q^{\frac{1}{2}}}
        \\&+ \dfrac{32LZ^2}{\sigma^2 QT\delta^2}
        +\dfrac{32L}{MQT\delta^2}
        +\dfrac{8L^3\lp d+6 \rp^3}{\lp d+4 \rp^3 T^2\sigma^2MQ}
        \\&+\dfrac{8\Bar{\omega}\sigma\lp d+4 \rp^{\frac{1}{2}}M^{\frac{1}{2}} Q^{\frac{1}{2}} L^{\frac{1}{2}}}{T^{\frac{1}{2}}},
    \end{split}
\end{equation} where $\Bar{\ell}_t(x)=\frac{1}{N}\sum_{i=1}^N\ell^i_t(x),$ $\Bar{\omega}\vcentcolon=\sum_{t=1}^T\omega_t,$ $x^*_{T+1} = \min_{i\in\{1,...,N\}}\argmin_x\ell^i_{T+1}(x),$  $\Delta=\Bar{\ell}_1(x^{1:N}_1) -\Bar{\ell}_{T+1}(x_{T+1}^*),$ and $\E[$\makebox[2ex]{$\cdot$}$]$ denotes $\E_{z_{1:T}^{1:N}}[$\makebox[2ex]{$\cdot$}$]$.
Furthermore, the number of time steps $T$ to obtain a $\xi$-accurate first order solution is 
\begin{equation}\label{eq:complexity}
\begin{split}
        &T = \\ &\mathcal{O}\lp\dfrac{\sigma^2 dMQ\lp\Delta^2+\Bar{\omega}^2\rp+M\lp\sigma^2+Z^4\rp}{\xi^2}+\dfrac{L^{\frac{5}{3}}}{\xi^{\frac{2}{3}}}+\dfrac{1}{\delta^2\xi}\rp.
\end{split}
\end{equation}
\end{theorem}
\begin{proof}[Sketch of proof]
    The general outline of the proof is very similar to that of the single-agent case. We define and work with the perturbed quantity $\Tilde{x}^{1:N}_t \vcentcolon= x^{1:N}_t-\eta\Bar{e}_t,$ where $\Bar{e}_t \vcentcolon= \dfrac{1}{N}\sum_{i=1}^N e^i_t.$ Additionally, our global loss function in this scenario is $\Bar{\Tilde{\ell}}_t\left(x^{1:N}_t\right) = \dfrac{1}{N}\sum_{i=1}^N \Tilde{\ell}^i_t\left(x^{1:N}_t\right).$ Using Assumptions \ref{assump:Lsmooth} and \ref{assump:drift}, we obtain
    \begin{equation}
        \begin{split}
    \Bar{\ell}_{\mu,t+1}\left(\Tilde{x}^{1:N}_{t+1}\right) &\leq \Bar{\ell}_{\mu, t}\lp\xtIN_t\rp \\
    &- \eta \left\langle \Bar{\Tilde{g}}_{\mu,t}\lp\xIN_t\rp, \nabla \Bar{\ell}_{\mu,t}\lp\xtIN_t\rp\right\rangle \\
    &+ \dfrac{L\eta^2}{2} \left\lVert\Bar{\Tilde{g}}_{\mu,t}\lp\xIN_t\rp\right\rVert^2 + \omega_t,
        \end{split}
    \end{equation}
    where $\omega_t = \max\{w_t^1,...,w_t^N\}.$ Taking expectations and algebraic manipulations lead to the main inequality with four terms:
    \begin{equation}
        \begin{split}
             \underbrace{\dfrac{\eta}{2}\left\lVert\nabla \Bar{\ell}_{\mu,t}\lp\xIN_t\rp\right\rVert^2}_\text{Term I} \leq \underbrace{\lb\Bar{\ell}_{\mu, t}\lp\xtIN_t\rp - \Bar{\ell}_{\mu,t+1}\left(\Tilde{x}^{1:N}_{t+1}\right)\rb}_\text{Term II} \\ + \underbrace{\dfrac{L\eta^2}{2} \E_{u^{1:N}_t, z^{1:N}_t}\lb\left\lVert\Bar{\Tilde{g}}_{\mu,t}\lp\xIN_t\rp\right\rVert^2\rb}_\text{Term III} + \underbrace{\dfrac{L^2\eta^3}{2}\lVert \Bar{e}_t\rVert^2}_\text{Term IV} + \omega_t.
        \end{split}
    \end{equation}
    Term II may be upper-bounded by means of a telescoping sum. Term I may be lower-bounded and Terms III and IV upper-bounded by quantities involving $\mathbb{E}_{z_{1:T}^{1:N}}[\lVert\nabla\Bar{\ell}_t(x^{1:N}_t)\rVert^2],$ using assumptions \ref{assump:contractive}, \ref{assumption:noise} and \ref{assump:boundedgrad}. Rearranging this, inserting the values for $\eta$ and $\mu$ and introducing $\xi$ to obtain an expression for the time complexity lead directly to the result. The complete proof may be found in the Appendix.
\end{proof}

Much like in the single-agent analysis, we note that the dominant term in the complexity is independent of the compression ratio $\delta$. 


\section{Experimental Results} \label{sec:experimental-results}
In this section, we explore two applications of the proposed method to multi-agent target tracking under communication constraints. The first application deals with the main focus of the work, i.e., multi-agent target tracking. The second is an alternative view of the problem involving an area-coverage problem.
 Our code used for the experiments is available online with the simulation video \cite{supp}.
 
\subsection{Target Tracking}\label{subsec:target-tracking}
We begin with the application of our proposed \nameref{alg:fed-ef-zo-sgd} multi-agent target tracking scenario detailed in the previous sections. In all experiments, we instantiate a central server, $N$ agents $\{\mathcal{A}_i\}_{i=1}^N$ and $N$ sources $\{\mathcal{S}_i\}_{i=1}^N$. The initial location of each agent is chosen uniformly at random from $[-100, 100]^2$ and each source from $[200, 400]^2$. Hence, $d=2$, i.e., we consider the target tracking problem on a $2-$dimensional plane, which is reasonable for the motivating example of delivery robots. Also, we instantiate the agents and sources in two separate clusters, with some initial distance between them. Each agent $\mathcal{A}_i$ aims to track source $\mathcal{S}_i$ and each $\mathcal{S}_i$ actively evades its tracker with maximum speed. This setting generalizes that of \cite{OPZO} to a scenario with multiple agents and sources. We use $x^i_{t}$, $z^i_{t}$ to denote the positions of $\mathcal{A}_i$ and $\mathcal{S}_i$ at time step $t$. Each $\mathcal{S}_i$ aims to maximize its distance to $\mathcal{A}_i$, by setting its velocity at each step to $\zeta^i_{t} = \beta(z^i_{t}-x^i_{t})/\lVert z^i_{t} - x^i_{t}\rVert$, i.e., moving directly away from $\mathcal{A}_i$ with speed $\beta=0.1$. 
\begin{figure}[h]
    \centering
    \includegraphics[width=2in]{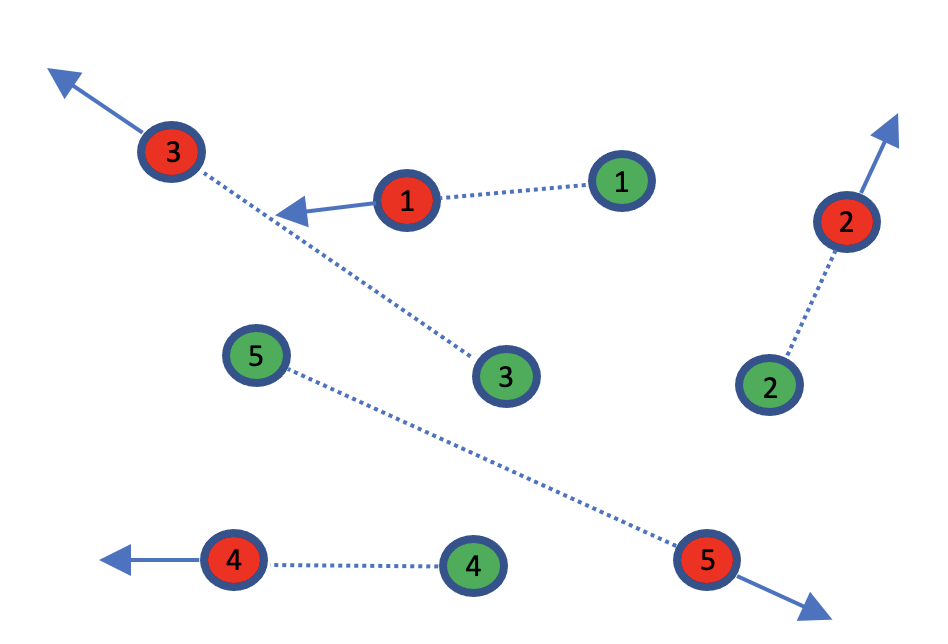}
    \caption{Illustration of 5 agents tracking 5 sources. Sources evade the agents by moving directly away from them.}
    \label{fig:agentvstarget}
\end{figure}An illustration of the movements of agents and sources is given in Fig. \ref{fig:agentvstarget}.

As explained in the previous sections, an additional contingency we introduce to the above setting is the requirement of a collision avoidance mechanism to prevent agents from unsafe maneuvers. To this end, we propose a two-level approach: \textbf{i)} on the local level within each agent, by means of local neighbor detection leveraging a judicious regularization term, and \textbf{ii)} coordination via the FL paradigm. With regard to \textbf{i)}, at every step of the simulation, we calculate the set of neighbors of each $\mathcal{A}_i$ as $D^i_t \coloneqq \{j \neq i:\lVert x_t^i-x_t^j\rVert\leq r\}$, where we set $r=10$. These neighbor sets determine the local loss function $\ell^i_t$ of $\mathcal{A}_i$ at time step $t$, which we define as:
\begin{equation}\label{localLoss}
    \ell_t^i(x_t^{1:N}, z^i_t) = \dfrac{1}{2}\lVert x_t^i - z_t^i \rVert^2 - \lambda\sum_{j\in D^i_t}\left(\lVert x_t^i-x_t^j\rVert^2-r^2\right),
\end{equation}
where $x^{1:N}_t = [(x^1_t)^T \cdots (x^N_t)^T]^T \in \mathbb{R}^{(Nd)}$ and $\lambda$ is the predetermined \textit{regularization parameter}. We note that the time-varying nature of these neighbor sets introduce time-variance to the loss functions, which is exactly the setting we examine in the theoretical analysis.
We divide the local loss function into two terms in order to simplify the notation in the subsequent calculation of the local ZO gradient estimator $g^i_{\mu,t}$:
\begin{equation}\label{divideLoss}
    \ell_t^i(x_t^{1:N}, z^i_t) = s_t^i(x_t^{i}, z^i_t) - \sum_{j\in D^i_t} r_t^{i,j}(x_t^i, x_t^j),
\end{equation}
where the loss due to source $s_t^i$ is given by $s_t^i(x_t^{i}, z^i_t) = \frac{1}{2}\lVert x_t^i - z_t^i \rVert^2$
and the loss due to regularization between agents $\mathcal{A}_i$ and $\mathcal{A}_j$, $r_t^{i,j}$, by $r_t^{i,j}(x_t^i, x_t^j) = \lambda(\lVert x_t^i-x_t^j\rVert^2-r^2)$. In terms of the scenario, one could see the regularization term as agents being able to sense other agents within a radius $r$ around its position. With regard to \textbf{ii)}, collision avoidance is ensured by means of federated aggregation of the local gradient estimators. The global loss function at $t$ is defined as $\Bar{\ell}_t(x_t^{1:N}, z_t^{1:N})=\frac{1}{N}\sum_{i=1}^{N}\ell_t^i(x_t^{1:N}, z^i_t)$, where $z^{1:N}_t$ is defined similarly to $x^{1:N}_t$.

\begin{figure*}[h!]%
    \centering
    \subfloat[\centering Tracking error vs iterations\label{fig:a}]{{\includegraphics[width=3in]{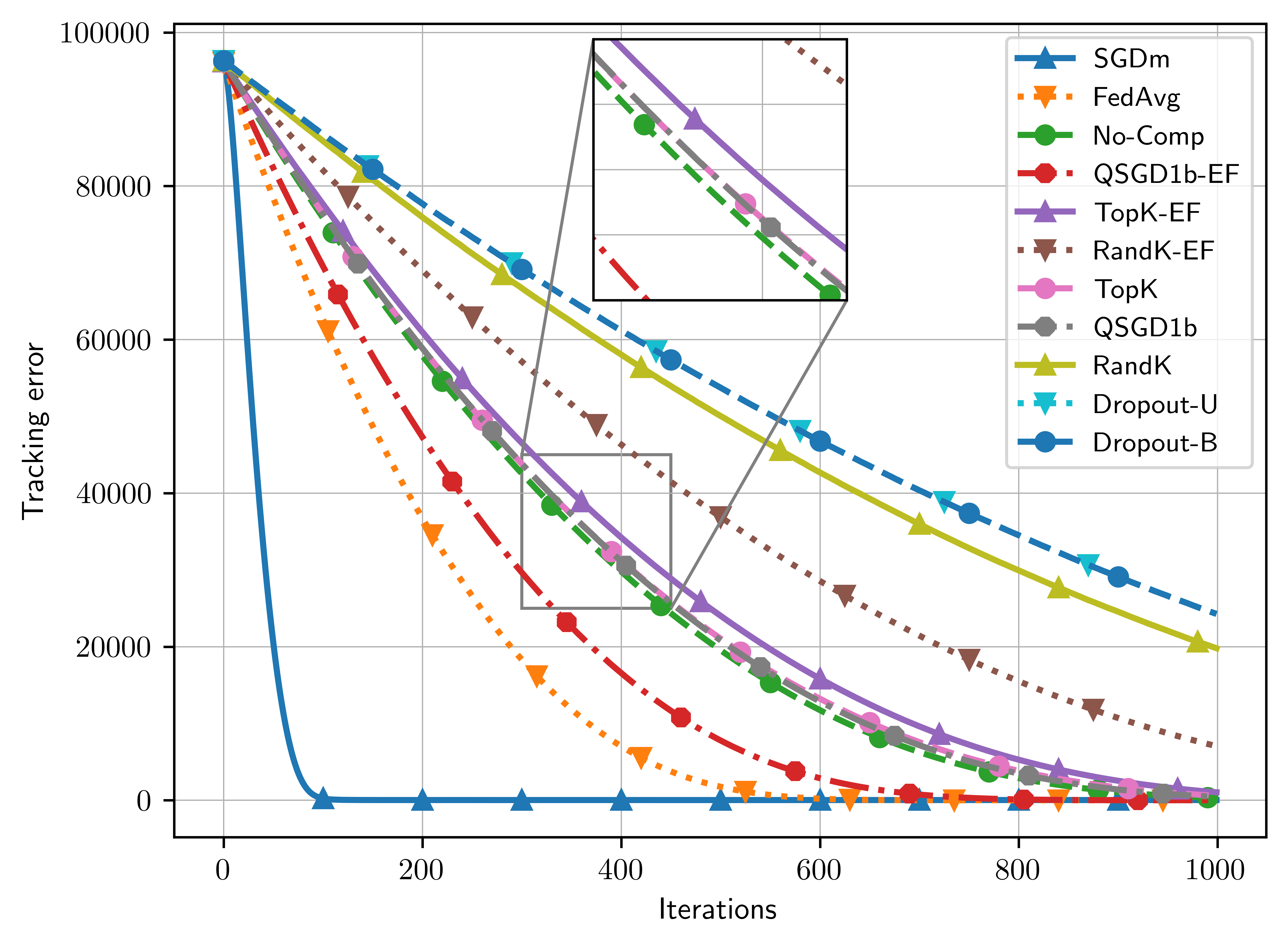} }}%
    \quad
    \subfloat[\centering Number of collisions vs iterations]{{\includegraphics[width=3in]{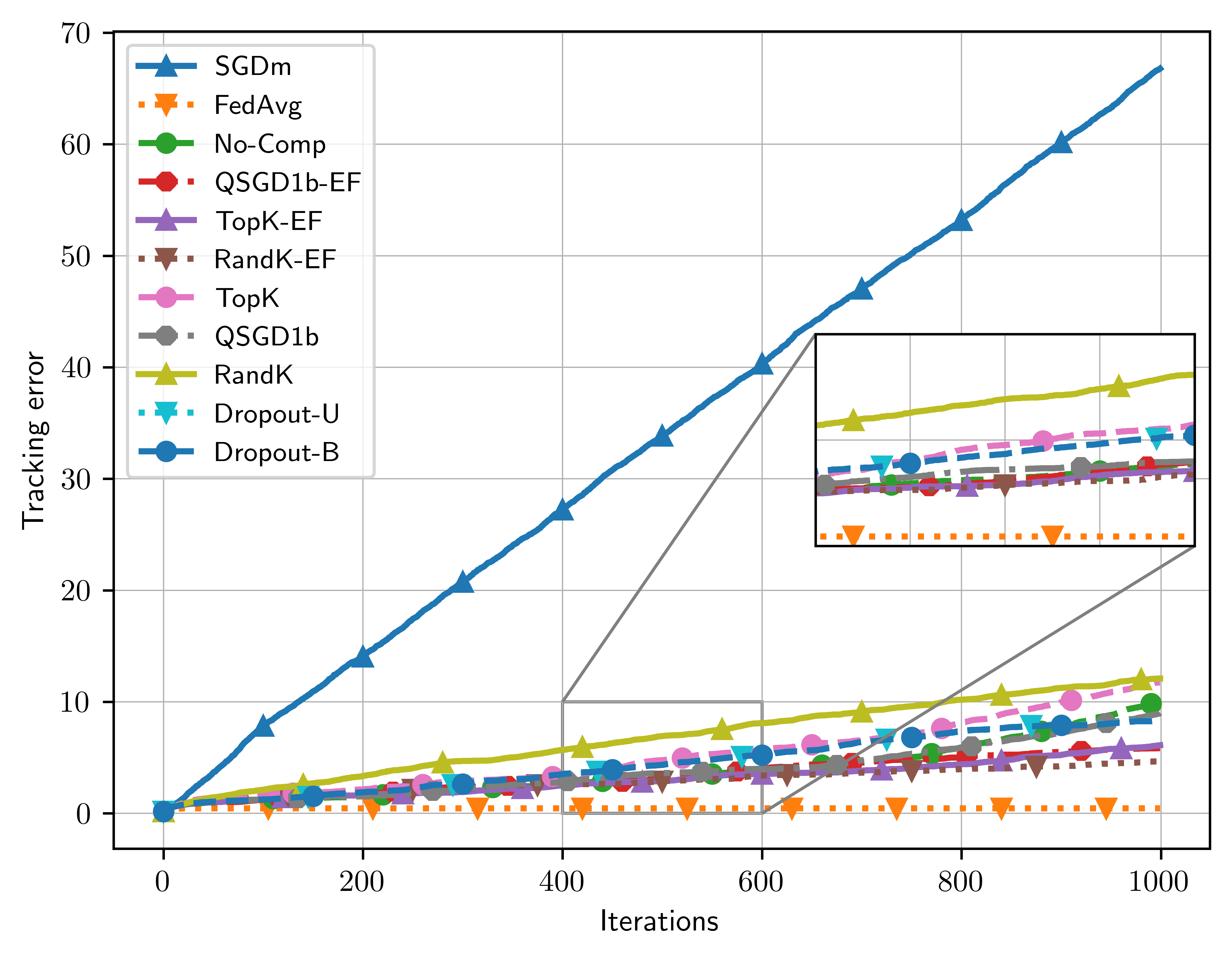} }}%
    \quad
    \subfloat[\centering Effect of varying $N$ on tracking error]{{\includegraphics[width=3in]{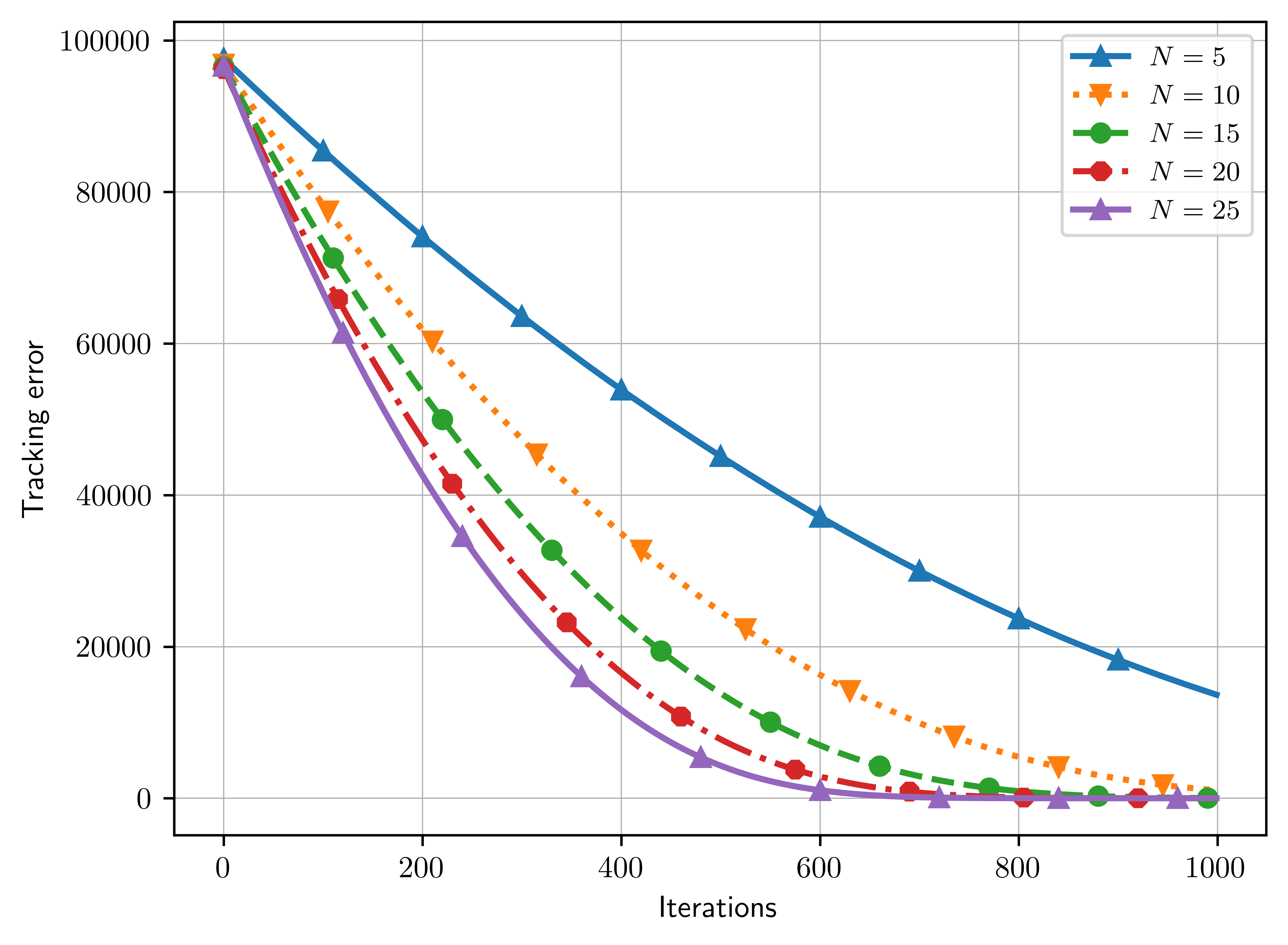} }}%
    \quad
    \subfloat[\centering Effect of varying $\lambda$ on number of collisions]{{\includegraphics[width=3in]{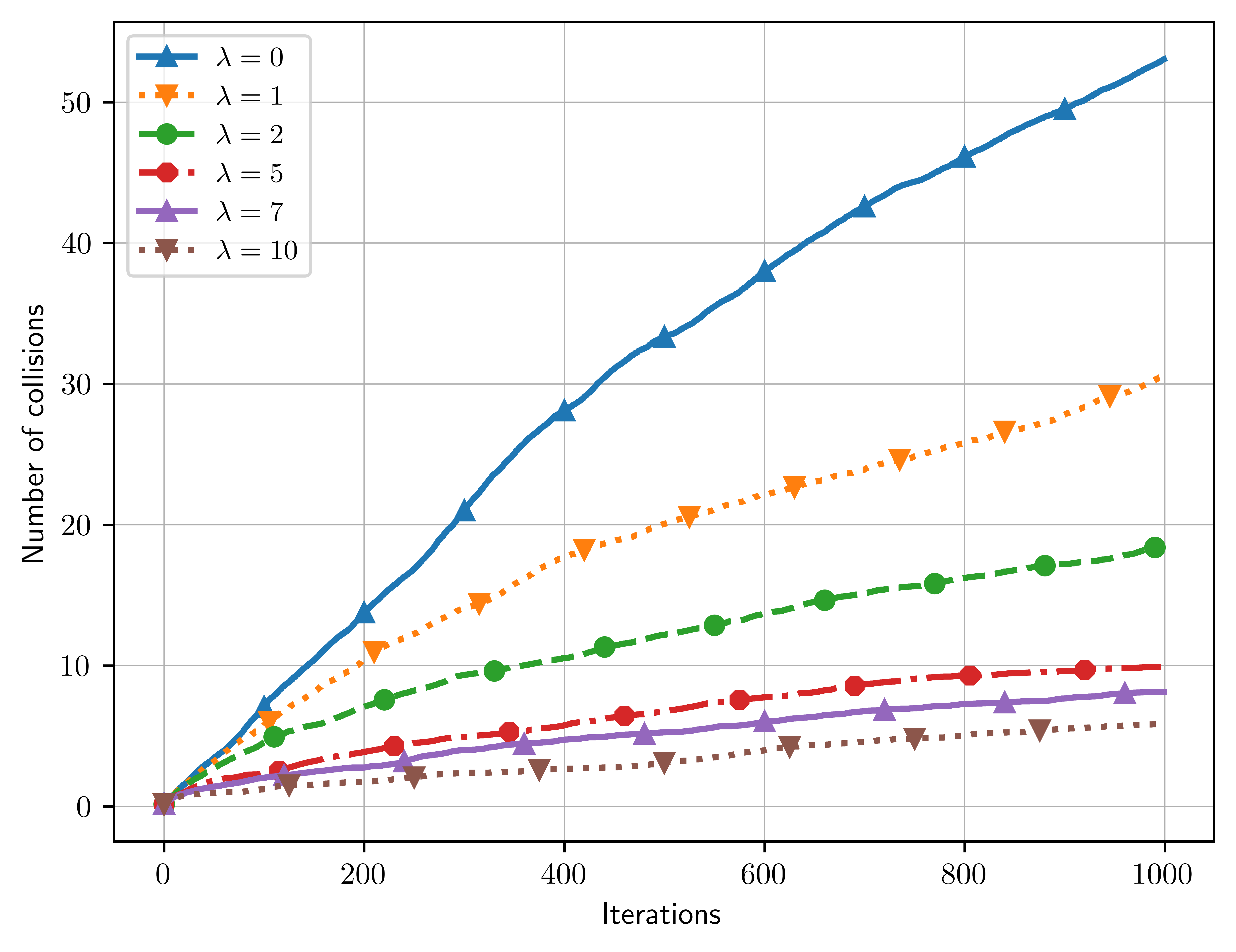} }}%
    \caption{  Results of \nameref{alg:fed-ef-zo-sgd} via the proposed scheme: (a) shows the average tracking error over $100$ runs of the simulation with different compression schemes and EF combinations in the FL paradigm, with SGDm being the non-FL benchmark algorithm, in which SGD with momentum is run locally on each agent with no communication, and FedAvg with $1-$bit QSGD compression and error feedback term being the FL benchmark algorithm. The difference of this benchmark algorithm with \nameref{alg:fed-ef-zo-sgd} is the fact that FedAvg uses first-order information. (b) shows collision numbers for the same experiment. (c) shows the average tracking errors over $100$ runs of the simulation with varying number of agents $N$, using the best-performing model of QSGD1b-EF (1-bit QSGD with error feedback term). The learning rate $\eta$ is set proportionally to $\sqrt{N}$. (d) shows the average numbers of collisions over $100$ runs of the simulation for varying values of the regularization parameter $\lambda$, using the best-performing model of QSGD1b-EF.}%
    \label{fig:example}%
\end{figure*}
Defining the neighborhood of two agents $\mathcal{A}_i$ and $\mathcal{A}_j$ in the above manner results in a symmetric relation. 
 To make the setting more interesting, we also introduce the concept of \textit{neighbor dropout} which aims to capture practical considerations such as imperfection in communication links and sensing capabilities. At each $t$, if $\mathcal{A}_i$ is to be added to  $\mathcal{D}^j_t$, a random number $X$ is sampled from $U[0, 1]$. If $X > p$, $i$ is added to $\mathcal{D}_t^j$, otherwise, it is dropped out. This leads to a more realistic scenario and opens up room for more meaningful collaboration between agents by breaking the symmetricity of the relation. If, for example, $\mathcal{A}_i$ is a neighbor of $\mathcal{A}_j$ but fails to detect it, we would expect $\mathcal{A}_j$ to compensate for this. Or worse, if both $\mathcal{A}_i$ and $\mathcal{A}_j$ fail to detect each other, we would expect $\mathcal{A}_k$ such that $k \in \mathcal{D}_t^i \cap \mathcal{D}_t^j$ to compensate for these detection failures. With the local loss function defined in \eqref{localLoss}, every agent $\mathcal{A}_i$ calculates a ZO gradient estimator $g^i_{\mu,t}$. Following the setup in \cite{OPZO}, we slightly modify the computation of the ZO estimator, by introducing a small change in the argument of the first function evaluation. Let
\begin{equation}\label{lossPlus}
    \begin{split}
        \ell^i_{t^+}(x_t^{1:N}, z^i_t) \vcentcolon= \dfrac{1}{2} \lVert x^i_t + \mu u^{i,i}_t - 
        (z^i_t + 0.5\zeta^i_t) \rVert^2  \\ - \lambda \sum_{j \in \mathcal{D}^i_t}\left(\lVert x^i_t + \mu u^{i,j}_t - (x^j_t + 0.5\xi^j_t)\rVert^2 - r^2\right),
    \end{split}
\end{equation}
where $u^{i,j}_t$ for all $j \in \mathcal{D}_t^i$ are drawn from $\mathcal{N}(0,I_{d})$ at time $t$ and $\zeta^i_t, \xi^i_t$ denote the velocities of agent $\mathcal{A}_i$ and source $\mathcal{S}_i$ at time $t$, respectively. Similar to  $\ell_t^i$ and \eqref{divideLoss}, we divide $\ell^i_{t^+}$ into two terms:
\begin{equation}\label{divideLossPlus}
    \ell^i_{t^+}(x_t^{1:N}, z^i_t) = s_{t^+}^i(x_t^{i}, z^i_t) - \sum_{j \in \mathcal{D}^i_t} r_{t^+}^{i, j}(x_t^{i}, x_t^j)
\end{equation}
where  $r_{t^+}^{i, j}(x_t^{i}, x_t^j) = \lambda (\lVert x^i_t + \mu u^{i,j}_t - (x^j_t + 0.5\xi^j_t)\rVert^2 - r^2)$ and $s_{t^+}^i(x_t^{i}, z^i_t) = \frac{1}{2} \lVert x^i_t + \mu u^{i,i}_t - (z^i_t + 0.5\zeta^i_t) \rVert^2$.

Now, we define $g^i_{\mu, t} = [(g^{i, 1}_{\mu, t})^T \cdots (g^{i, N}_{\mu, t})^T]^T \in \mathbb{R}^{(Nd)}$ where
\begin{equation}\label{gradientEstimatorComponents}
    g^{i, j}_{\mu, t} = \begin{cases}
   \hspace{5pt}\dfrac{s_{t^+}^i(x_t^{i}, z^i_t) - s^i_t(x_t^{i}, z^i_t)}{\mu}u^{i,i}_t & j = i, \\[10pt]
   \hspace{5pt}-\dfrac{r_{t^+}^{i, j}(x_t^{i}, x_t^j, \xi^j_t) - r_t^{i,j}(x_t^i, x_t^j)}{\mu} u^{i,j}_t  &  j \in \mathcal{D}^i_t,  \\[10pt]
    \hspace{5pt}0 \in \mathbb{R}^d & \text{otherwise}.
   \end{cases}
\end{equation}
In practice, it usually holds that for any $i \in \{1, \ldots, N\}$, $\lvert \mathcal{D}_t^i\rvert \ll N$, which results in a sparse $g^i_{\mu, t}$. Each agent then transmits its local ZO gradient estimator $g^i_{\mu, t}$ to the server. In scenarios with compression, each agent applies compression before transmission and transmits $\mathcal{C}(g^i_{\mu, t}+e^i_t)$ (see step 10 in \nameref{alg:fed-ef-zo-sgd}). The possible compression schemes that are used in the experiments are the ones that are detailed in Section \ref{section2}.  The server collects all of the transmitted (and possibly compressed) local gradient estimators and averages them, producing the aggregated global gradient estimator $\mathcal{G}_t$ of $\Bar{\ell}_t$: $\mathcal{G}_t = \dfrac{1}{N} \sum_{i=1}^N \mathcal{C}(g^i_{\mu, t}+e^i_t)$. Then, to keep the speed of the agents bounded in order to maintain a practically plausible simulation, the server normalizes $\mathcal{G}_t$ and then computes its estimation to the optimal position of every agent by $x^{1:N}_{t+1} = x^{1:N}_t - \eta\mathcal{G}_t$ where $\eta$ is the learning rate. With this formulation, $\eta$ determines the speed of the agents in the practical sense, since $\lVert \mathcal{G}_t \rVert=1$, therefore it only plays a role in determining the directions of the agents. The subsequent positions of agents are transmitted to the agents, without compression, and agents move to these positions. This process is illustrated on Fig. \ref{fig:FLillustration}. To gauge the performance of the model with respect to the number of collisions, we keep track of the number of collisions between agents by checking whether the position of any two agents $\mathcal{A}_i$ and $\mathcal{A}_j$ are close in Euclidean norm, the measure of closeness depends on the radii of the agents in the simulation. In all experiments, we set the collision radius $R=3$, i.e., we increment the collision counter whenever $\lVert x^i_t - x^j_t\rVert \leq 3$ for any two agents $\mathcal{A}_i$ and $\mathcal{A}_j$ such that $i \neq j$.


We conduct 3 types of experiments and depict the results on the 4 plots of Fig. \ref{fig:example}: In Fig. \ref{fig:example} {(a)} and Fig. \ref{fig:example} {(b)} we test the \nameref{alg:fed-ef-zo-sgd} algorithms' performance in terms of loss and number of collisions with various compression schemes. Fig. \ref{fig:example} {(c)}  compares the convergence of \nameref{alg:fed-ef-zo-sgd} for different numbers of agents $N$ while scaling the learning rate in proportion with $\sqrt{N}$, since the application bears theoretical resemblance to mini-batch SGD. Fig. \ref{fig:example} {(d)} demonstrates the effect of varying the regularization parameter $\lambda$ on the number of collisions. Unless otherwise stated, the parameters used in the experiments are $K=0.5$ for TopK and RandK, $p=0.5$ for Dropout, $\eta=1$, $\beta=0.1$, $p_N=0.5$, $d=2$, $N=20$, $r=10$ and $steps=1000$. $100$ instances of the simulation are run for each experiment, with the same fixed random seeds across different methods. In the first experiment, we also run SGD with momentum (SGDm) locally on each agent with no communication, i.e., without the FL paradigm as a benchmark algorithm. Additionally, a benchmark algorithm within the FL paradigm, we also look at the performance of FedAvg with 1-bit QSGD and error feedback mechanism, its key difference from FED-EF-ZO-SGD being that it uses first-order information.
 
 As Fig. \ref{fig:example} (a) demonstrates, the variant that leverages EF along with the QSGD compression scheme with $1-$bit quantization (QSGD1b-EF)  enjoys the fastest convergence and even outperforms the setting with no compression (No-Comp). This might be explained by the inherent noise introduced by quantization helping convergence. TopK with error feedback (TopK-EF), $1-$bit QSGD without error feedback (QSGD1b) and TopK without error feedback (TopK) perform virtually on par with the no compression setting. It is interesting to note that TopK seems to slightly outperform TopK-EF. These are followed in performance by RandK with error feedback (RandK-EF), and then RandK without error feedback (RandK). These are finally followed by Unbiased Dropout (Dropout-U) and Biased Dropout (Dropout-B), which perform equally well, but with a large gap to the best performers. It is expected for RandK-EF, RandK, Dropout-U and Dropout-B to take longer to converge, due to the high compression error that they inject in the communicated gradient estimators. Although, it appears that the error feedback helps the convergence of RandK significantly. We note that all of QSGD1b-EF, No-Comp, QSGD1b, TopK and TopK-EF converge within $1000$ iterations, with RandK-EF also coming very close. The non-FL benchmark algorithm SGDm outperforms all FL-based methods in terms of iterations needed for convergence, however the rate of convergence appears to be of the same order, and the results are comparable. The first-order FL benchmark algorithm FO-QSGD1b-EF enjoys slightly faster convergence than \nameref{alg:fed-ef-zo-sgd}, but the performance difference is marginal.

\begin{figure*}[h]%
    \centering
    \subfloat[\centering Effect of varying $\delta$ on tracking error with Dropout-B compression with error feedback term]{{\includegraphics[width=2in]{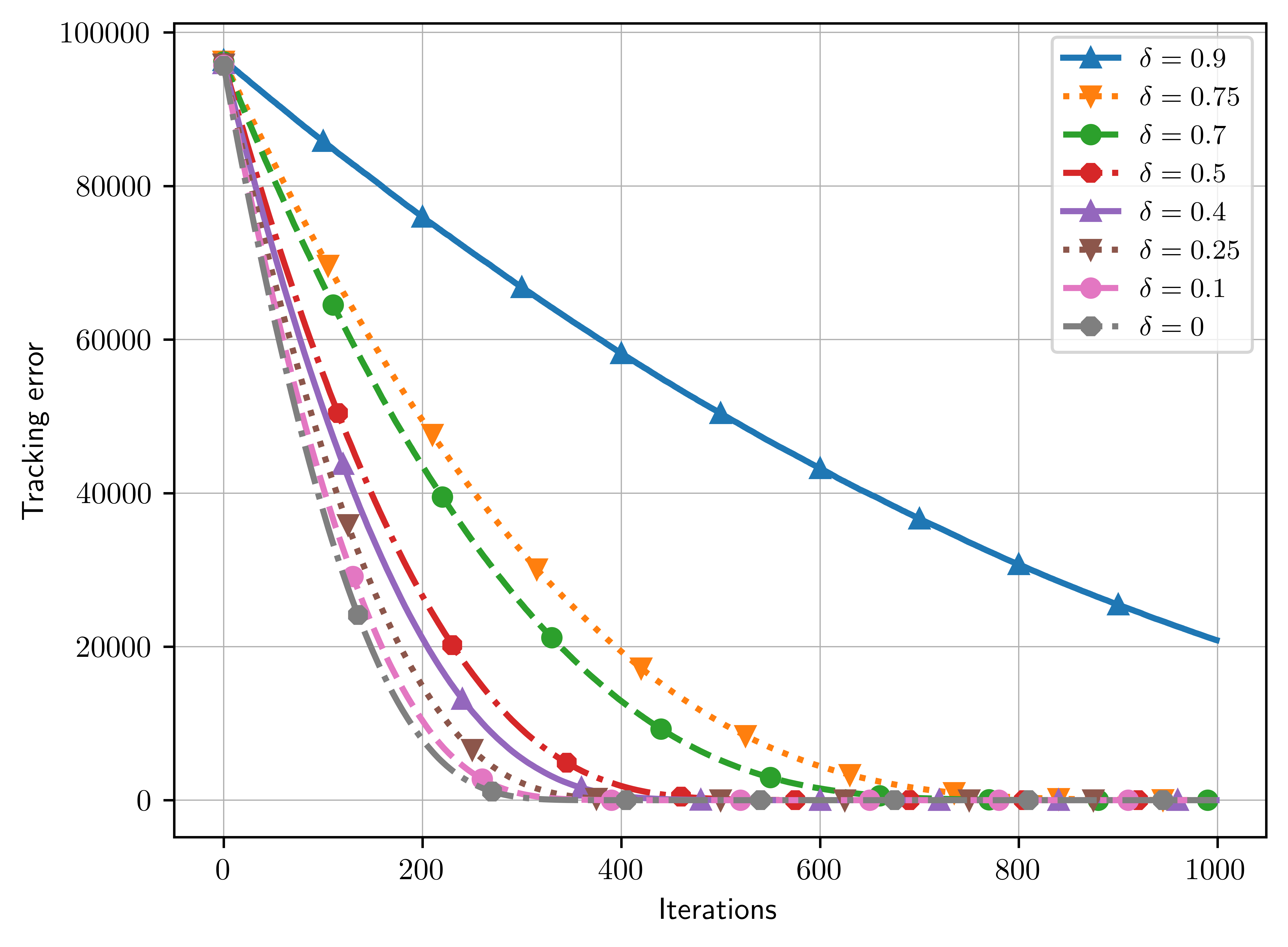} }}%
    \quad
    \subfloat[\centering Effect of varying number of bits in QSGD compression with error feedback term on tracking error]{{\includegraphics[width=2in]{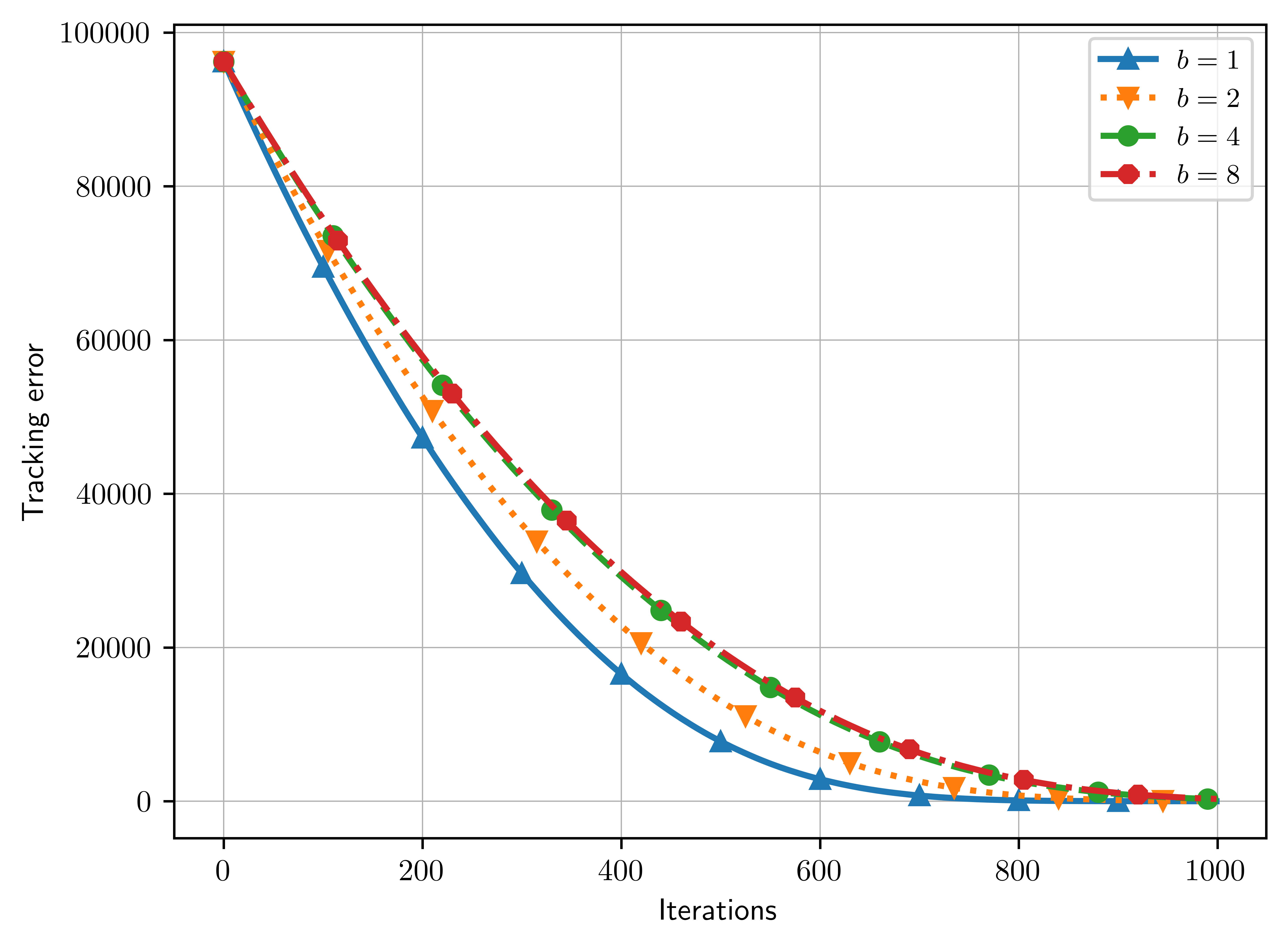} }}%
    \caption{Convergence results of \nameref{alg:fed-ef-zo-sgd} under different compression rates: (a) shows the tracking error over $100$ runs of the simulation with different values for $\delta$ under the Dropout-B compression scheme with error feedback term. (b) shows the tracking error over $100$ runs of the simulation with different numbers of bits used for the QSGD compression scheme with error feedback term.}%
\label{fig:delta-bits}
 \end{figure*}
 To evaluate the effectiveness of collaboration, we compare the number of collisions vs iterations for the same experiment in Fig. \ref{fig:example} (b). The results show that all of our FL-based methods far outperform the non-FL benchmark method of SGDm in terms of number of collisions. SGDm, which has no regard for collision prevention causes on average about $70$ collisions whereas all of our schemes, even the ones that do not achieve good convergence results such as RandK and Dropout-B cause at most about $10$ collisions on the average. This demonstrates the efficiency of the proposed \textit{regularization term.} 
\begin{figure}[h]
    \centering
    \includegraphics[width=2.5in]{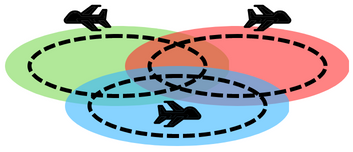}
    \caption{Illustration of the agents in the area coverage experiment. Each agent has the main objective of patrolling its designated area, following a circular route (indicated with the dashed curves). However, there is overlap between the areas, and the secondary objective is to discourage the agents from moving towards area that is already covered by another agent.}
    \label{fig:trajectory}
\end{figure}\noindent In Fig. \ref{fig:example} (c), we show the results of the second experiment, where we use the best-performing scheme in the first experiment, (QSDGD1b-EF) and test the convergence results with varying numbers of agents $N$. We make the observation that increasing the number of agents in the described multi-agent scenario is akin to increasing the batch size in mini-batch SGD, as by aggregating the messages received from agents, the server performs an update on the global objective. Thus, motivated by the theoretical studies of mini-batch SGD (see, e.g., \cite{SGDAnalysis}), to see the effect of varying the number of agents, we set the $\eta$ parameter proportional to $\sqrt{N}$. The values we use for $N$ are $5, 10, 15, 20,$ and $25$, with the respective $\eta$ values being $0.5, 0.71, 0.87, 1,$ and $1.12$. The main lines in the plot show the average tracking errors averaged over $100$ runs of the simulation. It can be seen that the comparison to mini-batch SGD may be justified, as the model converges roughly around the same iteration for all $N$ values except for $5$, when $\eta$ values are set proportionally.

In Fig. \ref{fig:delta-bits}, we demonstrate the effect of the compression parameter $\delta$ on the convergence of the tracking error. Although the theoretical analysis shows that the dominant term in the convergence bound is independent from the compression parameter $\delta$, the transient behavior of the convergence still depends on $\delta$. This is reflected in the experimental results. In Fig. \ref{fig:delta-bits} (a), we consider the effect of varying $\delta$ in the Dropout-B compression scheme. Here, $\delta$ corresponds to the probability that a gradient component will be dropped. The case of $\delta=0$ corresponds to when there is no compression. In these experiments, we set the step size $\eta=3$, to facilitate convergence in the highly compressed regime when $\delta=0.9$. It can be seen that even in the presence of extreme compression, convergence can be achieved by increasing the step size. Similarly, in In Fig. \ref{fig:delta-bits} (b), we consider the effect of varying the number of bits used in QSGD on the convergence of the tracking error. We simulate the experiment with number of bits $1, 2, 4,$ and $8$ and plot the results.
  

Finally, in Fig. \ref{fig:example} (d), we compare the effect of the regularization parameter $\lambda$ on the number of collisions. Similar to the second experiment, we use the best-performing scheme in the first experiment, QSDGD1b-EF. The values tested for $\lambda$ are $0, 1, 2, 5, 7,$ and $10$. We observe that, as expected, increasing the $\lambda$ parameter has a significant effect on decreasing the number of collisions. In the $\lambda=0$ scenario, which practically corresponds to no communication with regards to collision prevention among the agents, we observe on the average up to more than $50$ collisions, similar to the numbers observed in the first experiment with the benchmark SGDm method. Even the small value of $\lambda=1$ drops the number of collisions on average by almost half. We observe a drop in the number of collisions with each increment in $\lambda$, with $\lambda=10$ achieving less than $5$ collisions on average. This is naturally expected, since as we increase $\lambda$, the agents are more severely penalized when they get close to each other; hence, they maintain a safe distance to ensure a lower collision likelihood. It is intuitively clear that this effect demonstrates a diminishing marginal gain effect, in that the decrease in the number of collisions beyond the value of $\lambda=5$ seems to slow down.

\subsection{Area Coverage}

For the second application, we consider a scenario where multiple agents patrol a designated area by following a fixed trajectory, illustrated in Fig. \ref{fig:trajectory}. The goal of each agent is to maintain maximum total area coverage by avoiding crossing into areas already covered by other agents, while generally maintaining its fixed trajectory. A motivating example might be one where the agents are UAVs carrying out a ground coverage task of their designated areas, where the areas overlap in certain regions. Ideally, to have the maximum amount of ground coverage at any given time, we would want to discourage a UAV from approaching an overlapping region of its designated coverage area if it is already being covered by another UAV, since employing multiple UAVs for covering the same area would reduce the total amount of area covered. We claim that this can be seen akin to the first experiment in the following manner: If we increase the $r$ parameter of the agents to a suitable value, the collision prevention mechanism works in the way that the agents try not to cross into territories that are already covered by other agents. Also, the trajectories of the agents in their patrolling area can be modelled as perpetually tracking a target that follows said trajectory. In the experiments, we investigate 3 scenarios: the central server assigns agents new locations under compressed gradients with error feedback and nonzero regularization term, the same scenario but with the regularization term set to $0$ (which corresponds to a scenario without communication), and a scenario with no central aggregation, where agents run SGDm locally. We model the intended coverage areas of each agent as a disk of radius $5$, with overlapping regions ranging between $10\%$ to $25\%$. We report the number of ``collisions'', which in this case represents the number of area violations between agents, and present them in Table 1. 

\begin{table}[h]
\centering
\resizebox{\columnwidth}{!}{
\begin{tabular}{|c|c|c|c|c|c|c|}
\hline
\textbf{N} & \textbf{SGDm} & \textbf{No-Comp} & \textbf{QSGD3b} & \textbf{TopK} & \textbf{Dropout-B} & \textbf{RandK} \\ \hline
2                         & 0.8          & 0.0                     & 0.0           & 0.0            & 0.0                     & 0.0           \\ \hline
3                         & 9.0          & 0.8                     & 0.2           & 1.2            & 0.6                     & 0.4           \\ \hline
4                         & 12.4         & 1.0                     & 2.59          & 0.8            & 0.2                     & 0.2           \\ \hline
\end{tabular}}
\caption{Average number of collisions over $5$ runs of the simulation for varying $N$ with methods SGDm without central server; \nameref{alg:fed-ef-zo-sgd} with QSGD3b, TopK, Dropout-B and RandK, and No-Comp.}
\end{table} \noindent In the experiments, in addition to the $3$-agent scenario illustrated in Fig. \ref{fig:trajectory}, we also test $2$ and $4$ agent cases. We run each experiment for $7000$ iterations, with values $\lambda=100$, $N=2,3,4$, and the rest of the parameters have the same values as in the first experiment of the target-tracking problem. Running the simulation for $7000$ iterations corresponds to about $4$ full cycles of the agents around their circular trajectory. Again, we observe that the number of collisions reduces significantly by our \nameref{alg:fed-ef-zo-sgd} algorithm and the results obtained using a compressed gradient with error feedback are very close to the case where no compression is used. In some cases, compression with error feedback leads to even better results. This can be explained similarly to before in that owing to compressed gradients, we inject more noise to the gradients, introducing randomness to the trajectories of the agents, which helps avoid collisions. 

\section{Conclusion}\label{conclusion}
In this study, we tackled a problem of distributed online optimization with communication limitations, where multiple agents collaborate to track targets in a federated learning setting, limited to only zeroth-order information. The communication from the agents to the server was assumed to be constrained, and we addressed this constraint by compressing the communicated information along with an error feedback term. Our analysis showed that in the single-agent scenario, after $\mathcal{O}(\frac{d\sigma^2}{\xi^2})$ steps in the dominant term, the \nameref{alg:ef-zo-sgd} algorithm will reach a $\xi$-accurate first-order solution. In the multi-agent scenario, the \nameref{alg:fed-ef-zo-sgd} algorithm will converge to a $\xi$-accurate first-order solution after $\mathcal{O}(\frac{\sigma^2 dMQ(\Delta^2+\Bar{\omega}^2)+M\sigma^2+Z^4)}{\xi^2})$ steps in the dominant term. The dominant term in these convergence results are independent of the compression ratio $\delta$. The convergence of the \nameref{alg:fed-ef-zo-sgd} algorithm was confirmed through simulations. 

\textcolor{black}{As future work, one can investigate the collision constraints of each agent from safe reinforcement learning where in addition to maximizing rewards, agents must satisfy some constraints. This framework can be incorporated into our setting and can be analyzed from an optimization perspective. Additionally, rather than doing simple averaging at the central server, our work can be extended to a personalized federated learning setting where the losses are minimized by considering one step further of each agent. Further avenues for research include the examination of how the adaptive tuning of step sizes and the regularization parameters might change the convergence analysis. We note that the tuning of the regularization parameter is very related to dual formulations and Lagrangian methods in the general functional constrained optimization context. Finally, following the intuition presented in the experimental section, the effect of the number of agents on the variance of the stochastic gradients of the local loss functions may be studied. In this manner, as in mini-batch SGD, one might discover that incorporating a factor of $\sqrt{N}$ in the selection of the step size might accelerate convergence in a multi-agent scenario with $N$ agents.}

\section{Appendix. Proofs} \label{appendix:proofs}
\subsection{Lemmas}
We state several lemmas \textcolor{black}{from \cite{first-orderBook}}, mainly related to the zeroth-order method, which will be used in the main proofs.
Suppose $f(x)\in C_L^{1,1}(\mathbb{R}^d)$. Then, the following hold: 
\begin{lemma}\label{lem1}
    $f_{\mu}(x)\in C_{L_\mu}^{1,1}(\mathbb{R}^d)$, where $L_{\mu}\leq L$ \cite{first-orderBook}.
\end{lemma}

\begin{lemma}\label{lem2}$f_{\mu}(x)$ has the following gradient with respect to $x$:
\begin{equation}
    \nabla f_{\mu}(x) = \dfrac{1}{(2\pi)^{d/2}}\int\frac{f(x+\mu u) - f(x)}{\mu}u e^{(-\frac{1}{2}\lVert u\rVert^2)}\mathrm{d}u,
\end{equation} where $u\sim \mathcal{N}(0, I_d)$ \cite{first-orderBook}.
\end{lemma}
\begin{lemma}\label{lem3} For any $x \in \mathbb{R}^d$, we have
\begin{equation}
    \lvert f_{\mu}(x)-f(x)\rvert \leq \frac{\mu^2 Ld}{2},
\end{equation}\cite{first-orderBook}.
\end{lemma}
\begin{lemma}\label{lem4}  For any $x \in \mathbb{R}^d$, we have
\begin{equation}
    \lVert\nabla f_{\mu}(x) - \nabla f(x)\rVert \leq \frac{\mu}{2}L(d+3)^{\frac{3}{2}}
\end{equation}\cite{first-orderBook}.
\end{lemma}
\begin{lemma}\label{lem5}  For any $x\in \mathbb{R}^d$, we have
\begin{equation}
    \mathbb{E}_u\left[\left\lVert g_\mu\left(x\right)\right\rVert^2\right] \leq \frac{\mu^2}{2}L^2(d+6)^3 
    + 2(d+4)\lVert\nabla f(x)\rVert^2,
\end{equation} where $u\sim \mathcal{N}(0, I_d)$ and $g_\mu(x) = \frac{f(x+\mu u) - f(x)}{\mu}u$ \cite{first-orderBook}.
\end{lemma}
\begin{lemma}(Young's inequality)\label{lem:young}
For any $x, y \in \mathbb{R}^d$ and $\lambda > 0$, we have
\begin{equation}
    \langle x, y \rangle \leq \dfrac{\rVert x\lVert^2}{2\lambda} + \dfrac{\lVert y\rVert^2\lambda}{2}
\end{equation}\cite{first-orderBook}.

\end{lemma}
\subsection{Proof of Theorem 1} 
\begin{proof}\label{proof:single}
We assume that $z_t \in \mathbb{R}^d$ are \textit{i.i.d.} random variables for all $t\in \mathbb{Z}^+$. Furthermore, we drop the superscript notation present in the assumptions, since $i$ is always $1$ for the single-agent case.
Let $\Tilde{x}_t$ be defined as follows (following the analysis in \cite{errorterm}):
\begin{equation}\label{eqn1}
    \Tilde{x}_{t} := x_t - \eta e_t.
\end{equation} From \nameref{alg:ef-zo-sgd}, we know that $e_{t+1} = p_t - \mathcal{C}(p_t)$ and $p_t = \Tilde{g}_{\mu, t}(x_t) + e_t$, so we can rewrite $\Tilde{x}_{t+1}$ as
\begin{equation}\label{equ1}
\begin{split}
    \Tilde{x}_{t+1} & = x_{t+1}-\eta p_t + \eta \mathcal{C}(p_t) \\
                    & = x_t - \eta\mathcal{C}(p_t)-\eta \Tilde{g}_{\mu, t}(x_t)-\eta e_t + \eta\mathcal{C}(p_t) \\ 
                    & = x_t - \eta e_t - \eta \Tilde{g}_{\mu, t}(x_t) \\ 
                    & = \Tilde{x}_t -\eta \Tilde{g}_{\mu, t}(x_t),
\end{split}
\end{equation} 
where $\Tilde{g}_{\mu, t}(x_t) = \frac{\Tilde{\ell}_t(x_t+\mu u_t)-\Tilde{\ell}_t(x_t)}{\mu}u_t$ and $u_t \sim \mathcal{N}(0, I_d)$.
By Assumption \ref{assump:Lsmooth}, we can write the following:
\begin{equation}\label{lsmoothness}
\begin{split}
    \ell_{\mu, t}(\Tilde{x}_{t+1}) \leq \ell_{\mu, t}(\Tilde{x}_t) &+ \langle\nabla\ell_{\mu, t}(\Tilde{x}_t),\Tilde{x}_{t+1}-\Tilde{x}_t\rangle \\ &+ \frac{L}{2}\lVert\Tilde{x}_{t+1}-\Tilde{x}_t\rVert^2.
\end{split}
\end{equation}
Now by Assumption \ref{assump:drift}, we get:
\begin{equation}\label{boundeddrift}
    \begin{split}
        \ell_{\mu, t+1}(\Tilde{x}_{t+1})\leq \ell_{\mu, t}(\Tilde{x}_t) &- \eta\langle \Tilde{g}_{\mu, t}(x_t), \nabla\ell_{\mu, t}(\Tilde{x}_t)\rangle \\ &+ \frac{L\eta^2}{2}\lVert \Tilde{g}_{\mu, t}(x_t)\rVert^2 + \omega_t.
    \end{split}
\end{equation} Since $\nabla\ell_{\mu, t}(x_t)=\mathbb{E}_{u_t, z_t}\left[\Tilde{g}_{\mu, t}(x_t)\right]$, taking the expectation of both sides with respect to $u_t$ and $z_t$, we have the following: 
\begin{equation}\label{eq12}
\begin{split}
    \mathbb{E}_{u_t, z_t}\left[\langle \Tilde{g}_{\mu, t}(x_t),\nabla\ell_{\mu, t}(\Tilde{x}_t)\rangle\right] = \langle\nabla\ell_{\mu, t}(x_t), \nabla\ell_{\mu, t}(\Tilde{x}_t)\rangle,
\end{split}
\end{equation} and
\begin{equation}
    \begin{split}
        \langle\nabla\ell_{\mu, t}(x_t), \nabla\ell_{\mu, t}(\Tilde{x}_t)\rangle &= \frac{1}{2}\lVert \nabla\ell_{\mu, t}(x_t)\rVert^2 + \frac{1}{2}\lVert \nabla\ell_{\mu, t}(\Tilde{x}_t)\rVert^2 \\ &- \frac{1}{2}\lVert\nabla\ell_{\mu, t}(x_t)-\nabla\ell_{\mu, t}(\Tilde{x}_t)\rVert^2.
    \end{split}
\end{equation}
In the last step, we use the fact that $2\langle a,b\rangle = \lVert a\rVert^2 + \lVert b\rVert^2 - \lVert a-b\rVert^2$. Inserting this into \eqref{boundeddrift}, we get:
\begin{equation}
\begin{split}
    \ell_{\mu, t+1}(\Tilde{x}_{t+1}) &\leq \ell_{\mu, t}(\Tilde{x}_{t}) - \frac{\eta}{2}\lVert \nabla\ell_{\mu, t}(x_t)\rVert^2 \\
    &- \frac{\eta}{2}\lVert \nabla\ell_{\mu, t}(\Tilde{x}_t)\rVert^2 + \frac{L^2\eta}{2}\lVert x_t- \Tilde{x}_t\rVert^2\\
    &+ \frac{L\eta^2}{2}\mathbb{E}_{u_t, z_t}\left[\lVert \Tilde{g}_{\mu, t}(x_t)\rVert^2\right] + \omega_t.
\end{split}
\end{equation} Note that $\lVert\nabla\ell_{\mu, t}(x_t)-\nabla\ell_{\mu,t}(\Tilde{x}_t)\rVert^2 \leq L^2\lVert x_{t}-\Tilde{x}_t\rVert^2$ by Assumption \ref{assump:Lsmooth}, with subsequent application of Lemma \ref{lem1}. Also, we can drop -$\frac{\eta}{2}\lVert\nabla\ell_{\mu, t}(\Tilde{x}_t)\rVert^2$ because it is nonpositive. Using the fact that $\Tilde{x}_t - x_t = \eta e_t$, we get the main inequality:
\begin{equation}\label{equ13}
    \begin{split}
        \underbrace{\frac{\eta}{2}\lVert\nabla\ell_{\mu, t}(x_t)\rVert^2}_\text{Term I} &\leq \underbrace{\left[\ell_{\mu, t}(\Tilde{x}_t)-\ell_{\mu, t+1}(\Tilde{x}_{t+1})\right]}_\text{Term II} \\ &+ \underbrace{\frac{L\eta^2}{2}\mathbb{E}_{u_t,z_t}\left[\lVert \Tilde{g}_{\mu, t}(x_t)\rVert^2\right]}_\text{Term III} \\ &+ \underbrace{\frac{L^2\eta^3}{2}\lVert e_t\rVert^2}_\text{Term IV} + \omega_t.
    \end{split}
\end{equation} We will put an upper bound to the Terms II, III and IV and a lower bound to Term I. Starting with \textbf{Term III}, by Lemma \ref{lem5}, we know that
\begin{equation}\label{equ_term1}
\begin{split}
    \mathbb{E}_{u_t, z_{1:T}}\left[\lVert \Tilde{g}_{\mu, t}(x_t)\rVert^2\right] & \leq 2(d+4)\mathbb{E}_{z_{1:T}}\left[\lVert \Tilde{\nabla}\ell_t(x_t)\rVert^2\right] \\ &+ \frac{\mu^2L^2}{2}(d+6)^3,
\end{split}
\end{equation} where $\mathbb{E}_{z_{1:T}}[\lVert \Tilde{\nabla}\ell_t(x_t)\rVert^2]\leq M\mathbb{E}_{z_{1:T}}\left[\lVert \nabla\ell_t(x_t)\rVert^2\right] + \sigma^2$ by Assumption \ref{assumption:noise}. Note that, in this step, we use the the principle of causality and the fact that $z_t$ are \textit{i.i.d.} random variables. We can put the following upper bound to \textbf{Term II} by means of a telescoping sum and subsequent application of Lemma \ref{lem3}:
\begin{equation}\label{equ_term2}
\begin{split}
    \sum_{t=1}^T \left[\ell_{\mu, t}(\Tilde{x}_t) - \ell_{\mu, t+1}(\Tilde{x}_{t+1})\right] &= 
    \ell_{\mu, 1}(\Tilde{x}_1) - \ell_{\mu, T+1}(\Tilde{x}_{T+1}), 
\end{split}
\end{equation} 
and
\begin{equation}
    \begin{split}
        \ell_{\mu, 1}(\Tilde{x}_1) - \ell_{\mu, T+1}(\Tilde{x}_{T+1}) &\leq \mu^2 Ld + \ell_1(\Tilde{x}_1) - \ell_{T+1}(\Tilde{x}_{T+1})\\
    & = \mu^2 Ld + \ell_1(x_1) - \ell_{T+1}(\Tilde{x}_{T+1}),
    \end{split}
\end{equation}where we use the fact that $\ell(x_1)=\ell_1(\Tilde{x}_1)$, since $\Tilde{x}_1 = x_1$ by definition. Then, we can do the following:
\begin{equation}\label{termIIbound}
\begin{alignedat}{2}    
    \sum_{t=1}^T \left[\ell_{\mu, t}(\Tilde{x}_t) - \ell_{\mu, t+1}(\Tilde{x}_{t+1})\right] & \leq \mu^2 Ld + \ell_1(x_1) \\ &- \ell_{T+1}(\Tilde{x}_{T+1})\\
    & \leq \mu^2 Ld + \ell_1(x_1) \\ &- \ell_{T+1}(x^*_{T+1}),
\end{alignedat}
\end{equation} where $x_{T+1}^*\in\argmin_x\ell_{T+1}(x)$.
We can put the following lower bound to \textbf{Term I} by using Lemmas \ref{lem4} and \ref{lem:young}:
\begin{equation}\label{equ_termIII}
    \frac{1}{2}\lVert\nabla\ell_t(x_t)\rVert^2 - \frac{\mu^2L^2}{4}(d+3)^3 \leq \lVert\nabla\ell_{\mu, t}(x_t)\rVert^2.
\end{equation} Lastly, we can put the following upper bound to \textbf{Term IV} by Assumption \ref{assump:contractive} and Lemma \ref{lem:young}. (Due to space considerations, in the remainder of the proof, we denote the total expectation $\E_{u_{1:T}, z_{1:T}, \mathcal{C}_{1:T}}[$\makebox[2ex]{$\cdot$}$]$ as $\E[$\makebox[2ex]{$\cdot$}$]$.)
\begin{equation}\label{equ_termIV}
\begin{split}
    \E\left[\lVert e_{t+1}\rVert^2\right] & = \E\left[\lVert p_t - \mathcal{C}_{t}(p_t)\rVert^2\right]
    \\ &\leq (1-\delta)\E\left[\lVert p_t\rVert^2\right]\\ & = (1-\delta)\E\left[\lVert e_t +\Tilde{g} _{\mu, t}(x_t)\rVert^2\right] \\ & \leq (1-\delta)(1+\varphi)\E\left[\lVert e_t\rVert^2\right] + (1-\delta)\lp1+\frac{1}{\varphi}\rp\\ &\mathbb{E}_{u_{1:T}, z_{1:T}}\left[\lVert \Tilde{g}_{\mu, t}(x_t)\rVert^2\right],
\end{split}
\end{equation}
which we can write as,
\begin{equation}
\begin{split}
   \sum_{i=1}^t\left[(1-\delta)(1+\varphi)\right]^{t-i} (1-\delta)(1+\frac{1}{\varphi})\\\mathbb{E}_{u_{i}, z_{1:T}}\left[\lVert \Tilde{g}_{\mu, i}(x_i)\rVert^2\right], 
\end{split}
\end{equation}
for some $\varphi > 0$, $z_t, u_t, \mathcal{C}_t$ are \textit{i.i.d.}, and $\mathbb{E}_{\mathcal{C}_t}[$\makebox[2ex]{$\cdot$}$]$ denotes the expectation over the randomness at time $t$ due to the compression used. Note that by using Lemma \ref{lem5} and Assumption \ref{assumption:noise}, 
\begin{equation}
    \mathbb{E}_{u_{t}, z_{1:T}}[\lVert \Tilde{g}_{\mu, t}(x_t)\rVert^2] \leq A\mathbb{E}_{z_{1:T}}\left[\lVert\nabla\ell_t(x_t)\rVert^2\right] + B,
\end{equation}
where 
\begin{equation}\label{equ11}
\begin{split}
    & B = 2\sigma^2(d+4) + \frac{\mu^2L^2}{2}(d+6)^3 \; \text{and} \\
    & A = 2M(d+4).
\end{split}
\end{equation}
So we can rewrite \eqref{equ_termIV} as follows:
\begin{equation}
\begin{split}
        \mathbb{E}\left[\lVert e_{t+1}\rVert^2\right] \leq
        \sum_{i=1}^t\left[(1-\delta)(1+\varphi)\right]^{t-i}(1-\delta)(1+\frac{1}{\varphi}) \\ \left[A\mathbb{E}_{z_{1:T}}\left[\lVert \nabla\ell_i(x_i)\rVert^2\right]+B\right].
\end{split}
\end{equation}
If we set $\varphi\vcentcolon=\frac{\delta}{2(1-\delta)}$, then $1+\frac{1}{\varphi}\leq\frac{2}{\delta}$ and $(1-\delta)(1+\varphi)=(1-\frac{\delta}{2})$, so we get:
\begin{equation}\label{equ20}
\begin{split}
         \mathbb{E}\left[\lVert e_{t+1}\rVert^2\right] \leq \sum_{i=1}^t\left(1-\frac{\delta}{2}\right)^{t-i}\left[A\mathbb{E}_{z_{1:T}}\left[\lVert \nabla\ell_i(x_i)\rVert^2\right]+B\right] \\ \frac{2(1-\delta)}{\delta}.
\end{split}
\end{equation}
If we sum through all $ \mathbb{E}[\lVert e_t\rVert^2]$, we get:
\begin{equation}\label{equ21}
\begin{split}
    \sum_{t=1}^T \mathbb{E}\left[\lVert e_t\rVert^2\right] & \leq \sum_{t=1}^T\sum_{i=1}^{t-1}\left(1-\frac{\delta}{2}\right)^{t-i} \\ &\left[A\mathbb{E}_{z_{1:T}}\left[\lVert \nabla\ell_i(x_i)\rVert^2\right]+B\right]\frac{2(1-\delta)}{\delta}\\
    & \leq \sum_{t=1}^T\left[A\mathbb{E}_{z_{1:T}}\left[\lVert \nabla\ell_t(x_t)\rVert^2\right]+B\right] \\ &\sum_{i=0}^{\infty}\left(1-\frac{\delta}{2}\right)^i\frac{2(1-\delta)}{\delta} \\
    & \leq \sum_{t=1}^T\left[A\mathbb{E}_{z_{1:T}}\left[\lVert \nabla\ell_t(x_t)\rVert^2\right]+B\right]K,
\end{split}
\end{equation} where $K = \frac{2(1-\delta)}{\delta}\frac{2}{\delta}\leq\frac{4}{\delta^2}$. If we define $\Delta\vcentcolon=\ell_1(x_1) - \ell_{T+1}(x_{T+1}^*),$ where $x^*_{T+1}\in\argmin_x \ell_{T+1}(x),$ and combine the upper bounds derived in \eqref{equ_term1}, \eqref{equ_term2}, \eqref{equ_termIV}, and the lower bound derived in \eqref{equ_termIII} and insert them into \eqref{equ13}, we get the following:
\begin{equation}\label{eqn35}
\begin{split}
    &\sum_{t=1}^T  \frac{\eta}{4}\mathbb{E}_{z_{1:T}}\left[\lVert \nabla\ell_t(x_t)\rVert^2\right] - \frac{\eta\mu^2L^2}{8}(d+3)^3T \\ &\leq \mu^2 Ld + \Delta + \frac{T\mu^2L^3\eta^2}{4}(d+6)^3 + \frac{L\eta^2}{2}\sigma^2T 2(d+4) \\ &+ \frac{L\eta^2}{2}\times2M(d+4)\sum_{t=1}^T\mathbb{E}_{z_{1:T}}\left[\lVert\nabla\ell_t(x_t)\rVert^2\right] + \frac{\eta^3 L^2}{2} \\ &\times \frac{4}{\delta^2}T\left[2\sigma^2(d+4) + \frac{\mu^2L^2}{2}(d+6)^3\right]
     + \frac{\eta^3 L^2}{2} \\ &\times\frac{4}{\delta^2}\sum_{t=1}^T2M(d+4)\mathbb{E}_{z_{1:T}}\left[\lVert\nabla\ell_t(x_t)\rVert^2\right] + \sum_{t=1}^T\omega_t.
\end{split}
\end{equation} 
Now, since $z_t$'s are \textit{i.i.d.} for all $t\in \mathbb{Z}^+$, we have:
\begin{equation}
\begin{split}
    &\frac{E}{T}\sum_{t=1}^T\mathbb{E}_{z_{1:T}}\left[\lVert\nabla\ell_t(x_t)\rVert^2\right]  \\ &\leq \frac{\mu^2 L d + \Delta}{T} + \frac{\eta^2 L^3\mu^2(d+6)^3}{4} + L\eta^2\sigma^2(d+4) \\ &+ \frac{\eta\mu^2 L^2(d+3)^3}{8}
     + \frac{\eta^3 L^2}{\delta^2}4\sigma^2(d+4) \\&+ \frac{\eta^3 L^2}{\delta^2}\mu^2 L^2(d+6)^3 + \frac{1}{T}\sum_{t=1}^T\omega_t,
\end{split}
\end{equation} where 
\begin{equation}
\begin{split}
    E & = \frac{\eta}{4} - LM\eta^2(d+4)-\frac{L^2\eta^3}{\delta^2} 4M(d+4) \\ 
    & = \eta\left[\frac{1}{4} - LM\eta(d+4)\left(1+\frac{4L\eta}{\delta^2}\right)\right].
\end{split}
\end{equation} If $\eta\leq\frac{1}{4L}$, the first upper bound will instead be:
\begin{equation}
    1 + \dfrac{4L\eta}{\delta^2} \leq 1 + \dfrac{1}{\delta^2} = \dfrac{\delta^2 + 1}{\delta^2} \leq \dfrac{2}{\delta^2}.
\end{equation} We proceed to find an $\eta$ such that 
\begin{equation}
    \dfrac{2}{\delta^2} LM\eta(d+4)\leq \frac{1}{8}.
\end{equation} Then, we get 
\begin{equation}
    \eta \leq \frac{\delta^2}{16LM(d+4)}, 
\end{equation} which implies $E\geq \frac{\eta}{8}$. Multiplying all terms in the bound by $\frac{8}{\eta}$, 
\begin{equation}\label{eq:last}
\begin{split}
    &\frac{1}{T}\sum_{t=1}^T\mathbb{E}_{z_{1:T}}\left[\lVert\nabla\ell_t(x_t)\rVert^2\right] \leq \frac{8\Delta}{(\eta T)} + \frac{8\mu^2 L d}{\eta T} \\ &+ 2\eta L^3\mu^2(d+6)^3 + 8L\eta\sigma^2(d+4) + \mu^2 L^2(d+3)^3 \\ &+ \frac{32\eta^2 L^2}{\delta^2}\sigma^2(d+4) + \frac{8\eta^2 L^4\mu^2(d+6)^3}{\delta^2} + \frac{8}{\eta T}\sum_{t=1}^T\omega_t.
\end{split}
\end{equation} Let
\begin{equation}
    \eta = \frac{1}{\sigma\sqrt{(d+4)MTL}} 
    \quad\text{and}\quad \mu=\dfrac{1}{(d+4)\sqrt{T}}.
\end{equation}
Putting these values into \eqref{eq:last}, we get \eqref{equ22} as follows:
\begin{equation}\label{lastequ}
\begin{split}
    &\frac{1}{T}\sum_{t=1}^T\mathbb{E}\lVert\nabla\ell_t(x_t)\rVert^2 \leq \frac{8\Delta\sigma(d+4)^{\frac{1}{2}}M^{\frac{1}{2}}L^{\frac{1}{2}}}{T^{\frac{1}{2}}} \\
    &+ \frac{8\sigma dL^\frac{3}{2}M^\frac{1}{2}}{T^\frac{3}{2}(d+3)^\frac{3}{2}}
     + \frac{2(d+6)^\frac{3}{2}L^\frac{5}{2}}{\sigma (d+4)^\frac{5}{2}T^\frac{3}{2}M^\frac{1}{2}} + \frac{8\sigma(d+4)^\frac{1}{2}L^\frac{1}{2}}{M^\frac{1}{2}T^\frac{1}{2}} \\
     &+\frac{(d+3)^3L^2}{(d+2)^2T} 
     + \frac{32L}{\delta^2\sigma^2MT} + \frac{8(d+6)^3L^3}{\delta^2\sigma^2(d+4)^3MT^2} \\
    &+ \frac{8\Bar{\omega}\sigma(d+4)^\frac{1}{2}M^\frac{1}{2}L^\frac{1}{2}}{T^\frac{1}{2}}.
\end{split}
\end{equation}

Defining $\Bar{\omega}\vcentcolon=\sum_{t=1}^T\omega_t$, the number of times steps $T$ to obtain a $\xi$-accurate first order solution is 
\begin{equation}
    T = \mathcal{O}\left(\frac{d\sigma^2 L\Delta M}{\xi^2} + \frac{dL\Delta}{\delta^2\xi} + \frac{\Bar{\omega}\sigma^2dML}{\xi^2}\right).
\end{equation}
\end{proof}
\subsection{Proof of Theorem 2}
\begin{proof}\label{proof:multi}
    We assume in the following that $z^{1:N}_t \in \mathbb{R}^{Nd}$ are \textit{i.i.d.} random variables for all $t\in \mathbb{Z}^+$.
Similar to the analysis in the single-agent case, we begin by defining:
\begin{equation}
    \Bar{e}_t \vcentcolon= \dfrac{1}{N}\sum_{i=1}^N e^i_t,
\end{equation}
and
\begin{equation}
    \Tilde{x}^{1:N}_t \vcentcolon= x^{1:N}_t-\eta\Bar{e}_t.
\end{equation}
Additionally, our global loss function in this scenario is:
\begin{equation}
    \Bar{\Tilde{\ell}}_t\left(x^{1:N}_t\right) = \dfrac{1}{N}\sum_{i=1}^N \Tilde{\ell}^i_t\left(x^{1:N}_t\right).
\end{equation}
Now, we have:
\begin{equation}
\begin{split}
    \Tilde{x}^{1:N}_{t+1} &= x^{1:N}_{t+1}-\eta\Bar{e}_{t+1} \\
    &= x^{1:N}_{t+1} - \eta \dfrac{1}{N}\sum_{i=1}^N \left[ p^i_t - \mathcal{C}\left(p^i_t\right)\right] \\
    &= x^{1:N}_t - \eta \mathcal{G}_t - \eta\dfrac{1}{N}\sum_{i=1}^N \left[ p^i_t - \mathcal{C}\left(p^i_t\right)\right] \\
    &= x^{1:N}_t - \eta\dfrac{1}{N}\sum_{i=1}^N p^i_t \\
    &= x^{1:N}_t - \eta \dfrac{1}{N} \sum_{i=1}^N \left[\Tilde{g}^i_{\mu,t}\left(x^{1:N}_t\right) + e^i_t\right] \\
    &= \Tilde{x}^{1:N}_t - \eta \Bar{\Tilde{g}}_{\mu, t}\lp\xIN_t\rp,
\end{split}
\end{equation}
where we define $\Bar{\Tilde{g}}_{\mu, t}(\xIN_t) \vcentcolon=\frac{1}{N} \sum_{i=1}^N \Tilde{g}^i_{\mu,t}\left(x_t^{1:N}\right).$ Now, we have by Assumption \ref{assump:Lsmooth} that each $\ell^i_t$ is $L-$smooth, therefore, our global loss function $\Bar{\ell}_t$ is also $L-$smooth. Using Lemma \ref{lem1}, we write
\begin{equation}
\begin{split}
\Bar{\ell}_{\mu,t}\left(\Tilde{x}^{1:N}_{t+1}\right) \leq \Bar{\ell}_{\mu, t}\lp\xtIN_t\rp + \left\langle \nabla \Bar{\ell}_{\mu,t}\lp\xtIN_t\rp, \xtIN_{t+1} - \xtIN_t\right\rangle \\ + \dfrac{L}{2} \left\lVert \xtIN_{t+1} - \xtIN_t\right\rVert^2.
\end{split}    
\end{equation}
By Assumption \ref{assump:drift}, this implies
\begin{equation}\label{eq:multi-lsmooth-drift}
\begin{split}
\Bar{\ell}_{\mu,t+1}\left(\Tilde{x}^{1:N}_{t+1}\right) &\leq \Bar{\ell}_{\mu, t}\lp\xtIN_t\rp \\
&- \eta \left\langle \Bar{\Tilde{g}}_{\mu,t}\lp\xIN_t\rp, \nabla \Bar{\ell}_{\mu,t}\lp\xtIN_t\rp\right\rangle \\ &+\dfrac{L\eta^2}{2} \left\lVert\Bar{\Tilde{g}}_{\mu,t}\lp\xIN_t\rp\right\rVert^2 + \omega_t,
\end{split}
\end{equation}where $\omega_t = \max\{w_t^1,...,w_t^N\}.$
Now, since we have
\begin{equation}
    \begin{split}
        \E_{u^{1:N}_t}\lb\Bar{\Tilde{g}}_{\mu,t}\lp\xIN_t\rp\rb &= \E_{u^{1:N}_t}\lb\dfrac{1}{N} \sum_{i=1}^N \Tilde{g}^i_{\mu,t}\left(x^{1:N}_t\right)\rb \\ &= \dfrac{1}{N}\sum_{i=1}^N \nabla\Tilde{\ell}^i_{\mu,t}\lp\xIN_t\rp \\
        &= \nabla\Bar{\Tilde{\ell}}_{\mu,t}\lp\xIN_t\rp,
    \end{split}
\end{equation}
the following holds:
\begin{equation}
\begin{split}
    &\E_{u^{1:N}_t, z^{1:N}_t}\lb\left\langle\Bar{\Tilde{g}}_{\mu,t}\lp\xIN_t\rp, \nabla \Bar{\ell}_{\mu,t}\lp\xtIN_t\rp\right\rangle\rb \\ &=\left\langle\nabla \Bar{\ell}_{\mu,t}\lp\xIN_t\rp, \nabla \Bar{\ell}_{\mu,t}\lp\xtIN_t\rp\right\rangle =\dfrac{1}{2}\left\lVert\nabla \Bar{\ell}_{\mu,t}\lp\xIN_t\rp\right\rVert^2 \\ &+ \dfrac{1}{2}\left\lVert\nabla \Bar{\ell}_{\mu,t}\lp\xtIN_t\rp\right\rVert^2 
    \\&-\dfrac{1}{2}\left\lVert\nabla \Bar{\ell}_{\mu,t}\lp\xIN_t\rp - \nabla \Bar{\ell}_{\mu,t}\lp\xtIN_t\rp\right\rVert^2,
\end{split}
\end{equation}
since $\E_{z^{1:N}_t}[\nabla\Bar{\Tilde{\ell}}(x^{1:N}_t)] = \nabla\Bar{\ell}(x^{1:N}_t).$ Now, combining this with \eqref{eq:multi-lsmooth-drift} and using $L-$smoothness, we obtain:
\begin{equation}
\begin{split}
    \Bar{\ell}_{\mu,t+1}\left(\Tilde{x}^{1:N}_{t+1}\right) &\leq \Bar{\ell}_{\mu, t}\lp\xtIN_t\rp - \dfrac{\eta}{2}\left\lVert\nabla \Bar{\ell}_{\mu,t}\lp\xIN_t\rp\right\rVert^2 \\ &- \dfrac{\eta}{2}\left\lVert\nabla \Bar{\ell}_{\mu,t}\lp\xtIN_t\rp\right\rVert^2 
    \\&+ \dfrac{L^2\eta}{2}\left\lVert\xIN_t-\xtIN_t\right\rVert^2 \\ &+ \dfrac{L\eta^2}{2} \E_{u^{1:N}_t, z^{1:N}_t}\lb\left\lVert\Bar{\Tilde{g}}_{\mu,t}\lp\xIN_t\rp\right\rVert^2\rb + \omega_t
\end{split}
\end{equation} Note that the third term at the right-hand side of the inequality can be dropped because it is nonpositive. Using the definition of $\Tilde{x}^{1:N}_t$, and taking the expectation of both sides with respect to $u^{1:N}_t$ and $\zIN_t$, we have the following main inequality:
\begin{equation}\label{mainineq}
\begin{split}
        \underbrace{\dfrac{\eta}{2}\left\lVert\nabla \Bar{\ell}_{\mu,t}\lp\xIN_t\rp\right\rVert^2}_\text{Term I} &\leq \underbrace{\lb\Bar{\ell}_{\mu, t}\lp\xtIN_t\rp - \Bar{\ell}_{\mu,t+1}\left(\Tilde{x}^{1:N}_{t+1}\right)\rb}_\text{Term II} \\ &+ \underbrace{\dfrac{L\eta^2}{2} \E_{u^{1:N}_t, z^{1:N}_t}\lb\left\lVert\Bar{\Tilde{g}}_{\mu,t}\lp\xIN_t\rp\right\rVert^2\rb}_\text{Term III} \\ &+ \underbrace{\dfrac{L^2\eta^3}{2}\lVert \Bar{e}_t\rVert^2}_\text{Term IV} + \omega_t.
\end{split}
\end{equation} We will continue the proof by putting an upper bound to Terms II, III, and IV and a lower bound to Term I. Starting with \textbf{Term III}, using Jensen's inequality, we get
\begin{equation}\label{TermIstart}
\begin{split}
        &\E_{u^{1:N}_t, z^{1:N}_t}\lb\left\lVert\Bar{\Tilde{g}}_{\mu, t}(x^{1:N}_t)\right\rVert^2\rb \\ &= \E_{u^{1:N}_t, z^{1:N}_t}\lb\left\lVert \frac{1}{N}\sum_{i=1}^N \Tilde{g}^i_{\mu, t}(x^{1:N}_t)\right\rVert^2\rb \\ &\leq
    \frac{1}{N}\sum_{i=1}^N\E_{u^{1:N}_t, z^{1:N}_t}\lb\left\lVert \Tilde{g}^i_{\mu, t}(x^{1:N}_t)\right\rVert^2\rb.
\end{split}
\end{equation} Then, by Lemma \ref{lem5} we know
\begin{equation}\label{asd}
\begin{split}
    \mathbb{E}_{u^{1:N}_{1:T}, z^{1:N}_{1:T}}\left[\lVert \Tilde{g}^i_{\mu, t}(x^{1:N}_t)\rVert^2\right] & \leq 2(d+4)\\
    &\mathbb{E}_{z^{1:N}_{1:T}}\left[\lVert \nabla\Tilde{\ell}^i_t(x^{1:N}_t)\rVert^2\right] \\
    &+ \frac{\mu^2L^2}{2}(d+6)^3.
\end{split}
\end{equation}
\textcolor{black}{Using Assumption \ref{assumption:noise}, we have $\mathbb{E}_{z^{1:N}_{1:T}}[\lVert \nabla \Tilde{\ell}^i_t(x^{1:N}_t)\rVert^2] \leq M\mathbb{E}_{z^{1:N}_{1:T}}\left[\lVert \nabla\ell^i_t(x^{1:N}_t)\rVert^2\right] + \sigma^2$. Then, through application of Assumption \ref{assump:boundedgrad} and Lemma \ref{lem:young}, we have:}

\begin{equation}\label{}
\begin{split}
    \mathbb{E}_{u^{1:N}_{1:T}, z^{1:N}_{1:T}}\left[\lVert \Tilde{g}^i_{\mu, t}(x^{1:N}_t)\rVert^2\right] \leq 2(d+4)(MZ^2 + \sigma^2) \\ 
     + 2(d+4)MQ\mathbb{E}_{z_{1:T}^{1:N}}\left[\lVert\nabla\Bar{\ell}_t(x_t^{1:N})\rVert^2\right]\\
     + \frac{\mu^2L^2}{2}(d+6)^3.
\end{split}
\end{equation}



\noindent
For \textbf{Term II}, if we do a summation on both sides of \eqref{mainineq} from $t=1$ to $T$, we get a telescoping sum:
\begin{equation}\label{telescopian}
\begin{split}
        &\sum_{t=1}^T\left[\Bar{\ell}_{\mu, t}\lp\xtIN_t\rp - \Bar{\ell}_{\mu,t+1}\left(\Tilde{x}^{1:N}_{t+1}\right)\right] \\ &= \Bar{\ell}_{\mu, 1}\lp\xtIN_1\rp - \Bar{\ell}_{\mu,T+1}\left(\Tilde{x}^{1:N}_{T+1}\right).
\end{split}
\end{equation}By adding and subtracting $\Bar{\ell}_{1}(\Tilde{x}_1^{1:N})$ and $\Bar{\ell}_{T+1}(\Tilde{x}_{T+1}^{1:N})$ on both sides and using Lemma \ref{lem3}, we have:
\begin{equation}\label{TermIIstep2}
\begin{split}
    &\Bar{\ell}_{\mu, 1}\lp\xtIN_1\rp - \Bar{\ell}_{\mu,T+1}\left(\Tilde{x}^{1:N}_{T+1}\right) \\ &\leq \mu^2Ld + \Bar{\ell}_1(x^{1:N}_1) - \Bar{\ell}_{T+1}(\Tilde{x}^{1:N}_{T+1}). \\
    & \leq  \mu^2Ld + \Bar{\ell}_1(x^{1:N}_1) - \Bar{\ell}_{T+1}(x_{T+1}^*) \\
    & = \mu^2Ld + \Delta,
\end{split}
\end{equation} where $x^*_{T+1} = \min_{i\in\{1,...,N\}}\argmin_x\ell^i_{T+1}(x)$ and $\Delta=\Bar{\ell}_1(x^{1:N}_1) -\Bar{\ell}_{T+1}(x_{T+1}^*)$. Note that we use $\Tilde{x}^{1:N}_1 = x^{1:N}_1.$
For \textbf{Term I}, one should note that if $\ell^i_t(x)\in C^{1,1}_L$, then $\ell^i_{\mu, t}(x)\in C^{1,1}_L$ by Lemma \ref{lem1}. This implies that $\Bar{\ell}_{\mu, t}(x)\in C^{1,1}_L$ because $\Bar{\ell}_{\mu,t}(x)=\frac{1}{N}\sum_{i=1}^N\ell^i_{\mu,t}(x)$. Thus, using Lemmas \ref{lem4} and \ref{lem:young}, we get 
\begin{equation}\label{termIII}
    \frac{1}{2}\lVert \nabla\Bar{\ell}_t(x^{1:N}_t)\rVert^2 - \frac{\mu^2L^2(d+3)^2}{4} \leq \lVert\nabla\Bar{\ell}_{\mu, t}(x^{1:N}_t)\rVert^2. 
\end{equation}
Finally, for \textbf{Term IV}, we use the recursive summation similar to the one in the single-agent proof. We want to put an upper bound to $\lVert\Bar{e}_t\rVert^2$. We can do so by taking the expectation of both sides in \eqref{mainineq} with respect to $u^{1:N}_{1:T}, z^{1:N}_{1:T}, C_{1:T}$ and put an upper bound to $\mathbb{E}_{u^{1:N}_{1:T}, z^{1:N}_{1:T}, C_{1:T}}\left[\lVert \Bar{e}_t\rVert^2\right]$ instead. (Due to space considerations, in the remainder of the proof, we denote the total expectation $\E_{u^{1:N}_{1:T}, z^{1:N}_{1:T}, \mathcal{C}_{1:T}}[$\makebox[2ex]{$\cdot$}$]$ as $\E[$\makebox[2ex]{$\cdot$}$]$.) By Jensen's inequality, we can do the following:
\begin{equation}\label{TermIVJensen}
\begin{split}
    \mathbb{E}\left[\lVert \Bar{e}_t\rVert^2\right] = \mathbb{E}\left[\left\lVert \frac{1}{N}\sum_{i=1}^N e^i_{t}\right\rVert^2\right] &\leq \mathbb{E}\left[\frac{1}{N}\sum_{i=1}^N\left\lVert e^i_{t}\right\rVert^2\right]\\
    & = \frac{1}{N}\sum_{i=1}^N\mathbb{E}\left[\lVert e^i_{t}\rVert^2\right]
\end{split}
\end{equation}Note that putting an upper bound to the terms inside summation is nothing but putting an upper bound to the single-agent case, which we have done in Proof \ref{proof:single} of the single-agent setting. Hence, we know 
\begin{equation}\label{TermIIIsingleA}
\begin{split}
        \mathbb{E}\left[\lVert e^i_{t-1}\rVert^2\right] \leq \sum_{j=1}^{t-1}[(1-\delta)(1+\varphi)]^{t-1-j}(1-\delta)\left(1+\frac{1}{\varphi}\right) \\ \left[A\mathbb{E}_{z_{1:T}^{1:N}}\left[\lVert\nabla\ell^i_j(x^{1:N}_j)\rVert^2\right]+B\right].
\end{split}
\end{equation}Using this fact in \eqref{TermIVJensen}, we obtain
\begin{equation}
\begin{split}
        \mathbb{E}\left[\lVert e^{1:N}_{t}\rVert^2\right] \leq \frac{1}{N}\sum_{i=1}^N\sum_{j=1}^{t-1}[(1-\delta)(1+\varphi)]^{t-1-j} \\ (1-\delta)\left(1+\frac{1}{\varphi}\right) \left[A\mathbb{E}_{z_{1:T}^{1:N}}\left[\lVert\nabla\ell^i_j(x^{1:N}_j)\rVert^2\right]+B\right].
\end{split}
\end{equation} Using the same procedure in \eqref{equ21}, if we sum both sides through $t=1$ to $t=T$, we get the following inequality:
\begin{equation}\label{TermIVlastineq}
\begin{split}
        &\sum_{t=1}^T \mathbb{E}\left[\lVert e^{1:N}_{t}\rVert^2\right] \\  &\leq \frac{1}{N}\sum_{i=1}^N\sum_{t=1}^T\left[A\mathbb{E}_{z_{1:T}^{1:N}}\lVert \nabla\ell^i_t(x_t^{1:N})\rVert^2+B\right]K,
\end{split}
\end{equation}where $A=2M(d+4), B = 2\sigma^2(d+4)+\frac{\mu^2L^2(d+6)^3}{2}$ and $K = \frac{4(1-\delta)}{\delta^2}\leq\frac{4}{\delta^2}$. Another way of expressing \eqref{TermIVlastineq} is: 
\begin{equation}\label{TermIVFinalIneq}
\begin{split}
        &\sum_{t=1}^T \mathbb{E}\left[\lVert e^{1:N}_{t}\rVert^2\right] \\ &\leq \sum_{t=1}^T\left[A\left(\frac{1}{N}\sum_{i=1}^N\mathbb{E}_{z_{1:T}^{1:N}}\lb\lVert \nabla\ell^i_t(x_t^{1:N})\rVert^2\rb\right)+B\right]K.
\end{split}
\end{equation} 


\textcolor{black}{Using Assumption \ref{assump:boundedgrad}, we can write this as:}
\begin{equation}\label{termIVnewIneq}
\begin{split}
        &\sum_{t=1}^T \mathbb{E}\left[\lVert e^{1:N}_{t}\rVert^2\right] \\ &\leq \sum_{t=1}^T\left[A\left(Z^2 + Q\E_{z_{1:T}^{1:N}}\lb\lVert\nabla\Bar{\ell}_t(x^{1:N}_t)\rVert^2\rb\right)+B\right]K.    
\end{split}
\end{equation}
\textcolor{black}{If we now combine the upper bounds derived for Terms I, II and IV, and the lower bound derived for Term III and insert them into \eqref{mainineq}, we get the following inequality:}
\begin{equation}\label{newineq}
\begin{split}
    \frac{\eta}{4}\sum_{t=1}^T & \mathbb{E}_{z_{1:T}^{1:N}}\left[\lVert \nabla\Bar{\ell}_t(x^{1:N}_t)\rVert^2\right] - \frac{T\eta\mu^2L^2(d+3)^2}{8} \\
    & \leq \mu^2Ld + \Delta + TL\eta^2(d+4)(MZ^2 + \sigma^2)
    \\
    & + L\eta^2(d+4)MQ\left(\sum_{t=1}^T\mathbb{E}_{z_{1:T}^{1:N}}\lb\lVert\nabla\Bar{\ell}_t(x^{1:N}_t)\rVert^2\rb\right) \\
    & + \frac{TL^2\mu^2\eta^2(d+6)^3}{4} + \frac{2TL^2\eta^3K}{\delta^2} \\
    & + \frac{4L^2\eta^3(d+4)MQ}{\delta^2}\left(\sum_{t=1}^T\mathbb{E}_{z_{1:T}^{1:N}}\lb\lVert\nabla\Bar{\ell}_t(x^{1:N}_t)\rVert^2\rb\right)\\&+\sum_{t=1}^T\omega_t
\end{split}
\end{equation} where $K=2M(d+4)Z^2 + 2\sigma^2(d+4)+\frac{\mu^2L^2(d+6)^3}{2}$. After rearranging the terms and dividing both sides by $T$, we have the following inequality:

\begin{equation}\label{rearrange}
\begin{split}
    \frac{E}{T}\sum_{t=1}^T\E_{z_{1:T}^{1:N}}\lb\lVert\nabla\Bar{\ell}_t(x^{1:N}_t)\rVert^2\rb &\leq \frac{\mu^2Ld + \Delta}{T} \\&+ L\eta^2(d+4)(MZ^2 + \sigma^2) \\
    & + \frac{L^3\mu^2\eta^2(d+6)^3}{4} \\
    &+ \frac{2L^2\eta^3K}{\delta^2} + \dfrac{\Bar{\omega}}{T},
\end{split}
\end{equation}where $\Bar{\omega}\vcentcolon=\sum_{t=1}^T\omega_t$, and  
\begin{equation}
\begin{split}
    E & = \frac{\eta}{4} - LMQ\eta^2(d+4)-\frac{4L^2\eta^3MQ(d+4)}{\delta^2}  \\ 
    & = \eta\left[\frac{1}{4} - LM\eta(d+4)\left(1+\frac{4L\eta}{\delta^2}\right)\right].
\end{split}
\end{equation} If $\eta < \frac{1}{4L}$, the first upper bound will instead be:

\begin{equation}
    1 + \frac{4L\eta}{\delta^2}\leq 1 + \frac{1}{\delta^2} \leq \frac{2}{\delta^2}.
\end{equation} We proceed to find an $\eta$ such that 
\begin{equation}
    \frac{2L\eta(d+4)MQ}{\delta^2} \leq \frac{1}{8}.
\end{equation}Then, we get 
\begin{equation}
    \eta \leq \frac{\delta^2}{16LMQ(d+4)},
\end{equation}which implies $E\geq\frac{\eta}{8}$. Multiplying all the terms in the bound by $\frac{8}{\eta}$,
\begin{equation}
\begin{split}
    \frac{1}{T}&\sum_{t=1}^T\mathbb{E}_{z_{1:T}^{1:N}}\lb\lVert\nabla\Bar{\ell}_t(x^{1:N}_t)\rVert^2\rb \leq \frac{8\Delta}{\eta T} + \frac{8\mu^2Ld}{\eta T} \\
    &+ 8L\eta(d+4)(MZ^2 + \sigma^2) + 2L^3\mu^2\eta(d+6)^3 \\ &
    + \frac{32L^2\eta^2M(d+4)Z^2}{\delta^2} + \frac{32L^2\eta^2\sigma^2(d+4)}{\delta^2} \\
    &+ \frac{8L^4\mu^2\eta^2(d+6)^3}{\delta^2} + \dfrac{8\Bar{w}}{\eta T}.
\end{split}
\end{equation} Let 
\begin{equation}
\eta = \frac{1}{\sigma\sqrt{(d+4)MQTL}}\quad\mathrm{and}\quad\mu = \frac{1}{(d+4)\sqrt{T}}.
\end{equation}
Then, the number of times steps $T$ to obtain a $\xi$-accurate first order solution is:
\begin{equation}
\begin{split}
        &T = \\ &\mathcal{O}\lp\dfrac{\sigma^2 dMQ\lp\Delta^2+\Bar{\omega}^2\rp+M\lp\sigma^2+Z^4\rp}{\xi^2}+\dfrac{L^{\frac{5}{3}}}{\xi^{\frac{2}{3}}}+\dfrac{1}{\delta^2\xi}\rp.
\end{split}
\end{equation}

\end{proof}
\bibliographystyle{unsrt}
\renewcommand{\refname}{REFERENCES}
\bibliography{submission/ms}
\end{document}

%% file: defpack.tex
\usepackage{amsmath,graphicx,multicol}
\usepackage{amsfonts}
\usepackage{color,xcolor}
\usepackage{subcaption}
\usepackage{bbm}
\usepackage{cite}
\usepackage{amssymb}
\usepackage{tabularx,booktabs}
\usepackage{algorithm}
\usepackage{algcompatible}
\usepackage[utf8]{inputenc}
\usepackage[T1]{fontenc}
\usepackage[english]{babel}

\usepackage{amsthm}
\usepackage{bm}
\usepackage{url,hyperref}
\usepackage{cleveref}
\usepackage{mathtools}

\newtheorem{assumption}[]{Assumption}

\newtheorem{theorem}{Theorem}[]

\newtheorem{lemma}[]{Lemma}

\newcommand{\argmin}{\arg\!\min}

\DeclareMathOperator{\E}{\mathbb{E}}

\def\xIN{x^{1:N}}
\def\zIN{z^{1:N}}
\def\xtIN{\Tilde{x}^{1:N}}
\def\lp{\left(}
\def\rp{\right)}
\def\lb{\left[}
\def\rb{\right]}

\def\E{\mathbb{E}}

\def\act{\mathcal{A}}

\def\x{\mathbf{x}}

\newcommand{\RNum}[1]{\uppercase\expandafter{\romannumeral #1\relax}}
\makeatletter
\renewcommand{\fnum@figure}{Fig.~\thefigure}
\makeatother

